\documentclass[conference,10pt]{IEEEtran}
\IEEEoverridecommandlockouts

\usepackage{cite}
\usepackage{amsmath,amssymb,amsfonts}
\usepackage{algorithmic}
\usepackage{graphicx}
\usepackage{textcomp}
\usepackage{xcolor}
\usepackage{romannum}
\usepackage{multirow}
\usepackage{algorithm,algorithmic}
\usepackage{subfigure}
\usepackage{cleveref}
\usepackage{caption,setspace}
\usepackage{adjustbox}
\usepackage{caption}
\usepackage[font=footnotesize]{caption}
\newcommand{\qed}{\hfill $\square$}

\def\BibTeX{{\rm B\kern-.05em{\sc i\kern-.025em b}\kern-.08em
    T\kern-.1667em\lower.7ex\hbox{E}\kern-.125emX}}
\begin{document}

\title{LoLaFL: Low-Latency Federated Learning
via Forward-only Propagation\\

\thanks{
Received 19 December 2024; revised 9 June and 29 August 2025; accepted 22 October 2025.
A preliminary version of this work \cite{zhang2025lolafl} has been accepted by 2025 IEEE International Conference on Communications Workshops (ICC Workshops).

J. Zhang, J. Huang, and K. Huang are with the Department of Electrical and Electronic Engineering, The University of Hong Kong, Hong Kong. Emails: jrzhang@eee.hku.hk, jianhaoh@hku.hk, and huangkb@eee.hku.hk (Corresponding author: J. Huang and K. Huang).
}
}

\author{\IEEEauthorblockN{Jierui Zhang, \textit{Graduate Student Member, IEEE}, Jianhao Huang, \textit{Member, IEEE}, and Kaibin Huang, \textit{Fellow, IEEE}}
}

\maketitle

\begin{abstract}
Federated learning (FL) has emerged as a widely adopted paradigm for enabling edge learning with distributed data while ensuring data privacy. However, the traditional FL with deep neural networks trained via backpropagation can hardly meet the low-latency learning requirements in the sixth generation (6G) mobile networks. This challenge mainly arises from the high-dimensional model parameters to be transmitted and the numerous rounds of communication required for convergence due to the inherent randomness of the training process. To address this issue, we adopt the state-of-the-art principle of maximal coding rate reduction to learn linear discriminative features and extend the resultant white-box neural network into FL, yielding the novel framework of Low-Latency Federated Learning (LoLaFL) via forward-only propagation. LoLaFL enables layer-wise transmissions and aggregation with significantly fewer communication rounds, thereby considerably reducing latency. Additionally, we propose two \emph{nonlinear} aggregation schemes for LoLaFL. The first scheme is based on the proof that the optimal NN parameter aggregation in LoLaFL should be harmonic-mean-like. The second scheme further exploits the low-rank structures of the features and transmits the low-rank-approximated covariance matrices of features to achieve additional latency reduction. Theoretic analysis and experiments are conducted to evaluate the performance of LoLaFL. In comparison with traditional FL, the two nonlinear aggregation schemes for LoLaFL can achieve reductions in latency of over 87\% and 97\%, respectively, while maintaining comparable accuracies.
\end{abstract}


\section{Introduction}

With the growing volume of data and the increasing number of edge devices, the sixth generation (6G) mobile networks are envisioned to support a wide range of AI-based applications at the network edge, including augmented/mixed/virtual reality, connected robotics and autonomous systems, and smart cities and homes, among others \cite{letaief2019roadmap,saad2019vision,liu2025integrated}. 
To realize this vision, researchers have been motivated to develop technologies to deploy AI models at the network edge. These technologies, collectively called edge learning, leverage the mobile-edge-computing platform to train edge-AI models among edge servers and devices \cite{zhu2020toward,wang2024ultra}.
For its preservation of data privacy, federated learning (FL) emerges as a widely adopted solution for distributed edge learning, where local models are trained using local devices' data and sent to the server for updating the global model\cite{mcmahan2017communication,khan2021federated,li2024fedcir,lin2024efficient,chen2024fedmeld}. This collaborative training approach enables multiple devices and a server to train a global model without sharing raw data. However, FL faces its own challenges.
First, in scenarios where edge devices exhibit high mobility (e.g., autonomous cars and drones), they may move out of the range of an edge server before the learning process is completed.
Second, in contexts with dynamic environments and evolving user behaviors, timely model retraining is crucial.
These challenges necessitate the development of low-latency FL techniques to achieve faster response times \cite{singh2024towards,ma2019high,deng2022low,lim2020federated}.

However, achieving low-latency FL is challenging due to limited communication resources, which hinder the wireless exchange of high-dimensional stochastic gradients or models between devices and edge servers\cite{bennis2018ultrareliable,yang2021spectrum}. Researchers have explored various approaches to allocate network resources and schedule participating devices such as wireless power transfer \cite{zeng2021wirelessly}, resource allocation \cite{yang2020energy,zhang2023federated,wen2020joint,wang2024spectrum}, and client scheduling \cite{zeng2023federated,chen2020joint}, to improve task performance. For example, in \cite{chen2020joint}, a problem of joint learning, resource allocation, and user selection for FL over wireless networks was formulated and solved, improving the inference accuracy. Besides, the popular over-the-air computation (AirComp) technology is widely adopted to leverage the property of waveform superposition over a multi-access channel to realize simultaneous model uploading and over-the-air aggregation, thereby accelerating FL \cite{zhu2019broadband,yang2020federated,liu2023over}.

Despite these efforts to optimize the resource allocation for latency reduction, the bottleneck of low-latency FL lies in the high-dimensional gradients or model parameters to be transmitted and the numerous rounds for convergence \cite{chen2024federated,xia2021federated,lin2024split}.
For the first problem, approaches are considered to reduce the number of parameters to be transmitted. For example, model splitting introduces a method where the global model can be partitioned and distributed between server and devices for collaborative training, thereby reducing latency by transmitting only a portion of the gradients\cite{lin2024adaptsfl,ni2024fedsl}. Some lossy compression techniques can also be utilized. In particular, sparsification helps to drop insignificant model parameters \cite{zhang2023joint}, and quantization enables the use of fewer bits to represent an element for transmission \cite{zhu2020one}. For the second problem, techniques like federated transfer learning can be utilized for model initialization and speed up the convergence\cite{xu2022secure}.
In essence, these two problems arise from the fundamental nature of deep neural networks (DNNs), such as Convolutional Neural Networks (CNNs) \cite{lecun1998gradient}.
Specifically, their architectures are typically designed using a heuristic approach, and the training process involves random initialization and multiple rounds of weight updates via backpropagation (BP). This design principle, training method, and numerous heuristic techniques involved in DNNs, collectively earn them the label of \emph{black-box} \cite{rumelhart1986learning,he2015delving,xu2015empirical,srivastava2014dropout,montavon2018methods}.
The bottleneck cannot be easily overcome without challenging the current paradigm of FL, which necessitates the adoption of novel NN architectures and training approaches along with the design of a compatible FL framework.

Recently, the new approach of \emph{white-box} has emerged, which focuses on providing rigorous mathematical principles to understand the underlying mechanisms of both the architecture and parameters of DNNs\cite{ma2022principles}. One notable example is the recent work in \cite{chan2022redunet}, which proposes a \emph{forward-only} algorithm to directly construct an AI model from the intrinsic structures of data, without the need for multiple rounds of BP.
Taking the classification task as an example, many real-world datasets exhibit specific structures and distributions in high-dimensional spaces. Then the objective for white-box DNNs is to learn the intrinsic structures underlying the data, namely the linear discriminative features in order to achieve effective classification \cite{chan2022redunet,ma2022principles}. The principle of maximal coding rate reduction (MCR$^2$) was proposed in \cite{yu2020learning} to obtain these kinds of features from data. Therein, the so-called coding rate was introduced to quantify the volume of feature space spanned by the features up to a specific precision \cite{ma2007segmentation,cai2024multi}, as inspired by the classic rate-distortion theory \cite{cover1999elements}. MCR$^2$ calls for maximizing the volume of the entire feature space while minimizing that of the summed feature sub-spaces, which can be achieved through step-by-step feature transformation; the gradient information from each step forms the layer parameters of a novel \emph{white-box neural network} constructed in a \emph{forward-only} manner\cite{chan2022redunet}. More surprisingly, it has been demonstrated that the white-box neural network has a similar architecture and comparable task accuracy to its black-box counterpart (e.g., the well-known ResNet \cite{he2016deep}).

These white-box neural networks have two distinct characteristics. First, their parameters can be calculated from features directly and deterministically using formulae, eliminating the need of BP to update parameters. Second, such a model is constructed \emph{only forwardly}, with each layer obtained based on the information from the previous layer.
We advocate the design of low-latency FL by adopting the new training approach of white-box neural network to leverage its above characteristics. 
The first characteristic facilitates rapid convergence in model training. On the other hand, the second characteristic presents a new opportunity to advance low-latency FL: in each communication round, only the parameters of the latest layer instead of the whole model need to be transmitted.
However, how to apply this white-box approach to FL in order to achieve low-latency edge learning remains an open problem. Solving it requires designing unique and compatible techniques for parameter transmissions and aggregation.

To this end, this paper presents a novel \emph{low-latency federated learning} (LoLaFL) framework via forward-only propagation. LoLaFL features layer-wise transmissions and aggregation with much fewer communication rounds than traditional FL, thereby reducing communication latency dramatically. Specifically, in each communication round, only the latest layer targeted for updating in the round, rather than the entire NN, is uploaded for aggregation and subsequently updated with the received aggregated one. The key differences between LoLaFL and traditional FL are summarized in Table~\ref{LoLaFL vs traditional FL}.
The key contributions and findings of this paper are summarized as follows.

\renewcommand{\arraystretch}{1.25}
\begin{table*}[t]
\normalsize
\caption{Comparison between LoLaFL and Traditional FL}
\begin{adjustbox}{center}
\resizebox{1.9\columnwidth}{!}{\begin{tabular}{|c|cccccc|}
\hline
\multirow{2}{*}{\textbf{Framework}} & \multicolumn{6}{c|}{\textbf{Characteristics}}                                                                                                                                                                                                \\ \cline{2-7} 
                                    & \multicolumn{1}{c|}{\textbf{Nature}} & \multicolumn{1}{c|}{\textbf{Training}} & \multicolumn{1}{c|}{\textbf{Target}}      & \multicolumn{1}{l|}{\textbf{What to transmit}} & \multicolumn{1}{c|}{\textbf{Aggregation}} & \textbf{Latency} \\ \hline
Traditional FL                      & \multicolumn{1}{c|}{Black-box}       & \multicolumn{1}{c|}{Backpropagation}          & \multicolumn{1}{c|}{Minimized loss}      & \multicolumn{1}{c|}{Entire model}               & \multicolumn{1}{c|}{Linear}               & High             \\ \hline
LoLaFL                              & \multicolumn{1}{c|}{White-box}       & \multicolumn{1}{c|}{Forward-only}              & \multicolumn{1}{c|}{Discriminative features} & \multicolumn{1}{c|}{Single layer}                   & \multicolumn{1}{c|}{Nonlinear}           & Low              \\ \hline
\end{tabular}}
\end{adjustbox}
\label{LoLaFL vs traditional FL}
\end{table*}

\begin{itemize}
    \item \textbf{LoLaFL Framework}: The proposed framework consists of multi-round operations with the number of rounds determined by the number of model layers. In each communication round, the local parameters are calculated at edge devices based on local features and subsequently uploaded for aggregation at the edge server. The aggregated global parameters are then broadcast and used for local layer construction and local feature transformation. Unlike traditional FL, which requires the transmission of the entire model for updates at each communication round, LoLaFL only computes and transmits one layer of the neural network. This not only achieves low latency but also alleviates resource-constrained devices' computation load.
    \item \textbf{Nonlinear Aggregation}: 
    We have proved that the optimal aggregation for the global parameters of the white-box NN is not the traditional arithmetic mean (see e.g., FedAvg\cite{mcmahan2017communication}) but the harmonic mean (HM) of the local parameters. Motivated by the finding, we propose two nonlinear aggregation schemes for LoLaFL, which are more flexible and powerful in capturing the complex relationships between the local and global parameters.
    Furthermore, we devise a scheme for compressing uploaded parameters by leveraging the low-rank structures of features. Incorporating this scheme to enhance the HM-like aggregation results in further reduction on latency.
    \item \textbf{Performance Analysis}: First, the communication latency and computational complexity are derived theoretically, which are demonstrated to be primarily determined by the dataset. Specifically, the latency and complexity are found to be proportional to the square and cube of data's dimensionality, respectively, and both are proportional to the number of classes in the dataset. In contrast, the latency and complexity of traditional FL are primarily determined by the number of parameters and layers, with their effects being more significant when both the data dimensionality and the class number are small. Therefore, we conclude that LoLaFL exhibits smaller latency and complexity than traditional FL when both the data dimensionality and the class number are small. Next, we mathematically prove that the features or raw data cannot be recovered from the transmitted parameters, ensuring that LoLaFL is privacy-preserving.
    \item \textbf{Experiments}: The experiments are conducted in various scenarios to examine the performance of LoLaFL. By benchmarking against traditional FL, the results reveal that our two schemes for LoLaFL can achieve more than \emph{$87\%$ and $97\%$ reductions in latency}, respectively, while maintaining comparable accuracies. The convergence speed of LoLaFL is ten times faster than traditional FL in terms of communication round. Additionally, LoLaFL demonstrates greater robustness with non-IID data.
    \end{itemize}

The remainder of this paper is organized as follows. Section \ref{principles} compares the principles of black-box and white-box neural networks. The system model is introduced in Section \ref{system model}.
The LoLaFL framework and two novel nonlinear aggregation schemes for it are presented in Section \ref{LoLaFL}. In Section \ref{Theoretic Analysis}, the communication latency and computational complexity are analyzed and the privacy guarantee of LoLaFL is characterized. Experimental results are provided in Section \ref{simulation}, followed by concluding remarks in Section \ref{concluding remarks}.

\section{Principle Comparisons: Black-box versus White-box}\label{principles}

\subsection{Black-box DNNs via BP}\label{DNN as Black-box}

DNNs can be seen as a nonlinear function that maps inputs to their corresponding outputs. In practice, the common approach is to design a heuristic architecture, and choose a loss function to measure the discrepancy between the network outputs and expected outputs for a specific learning task. The process to minimize the loss function, known as training, typically involves initializing the network parameters and then updating them via BP \cite{rumelhart1986learning}.
For classification, the global loss function is given by
\begin{equation}
    F(\mathbf{w})=\frac{1}{\left\lvert \mathcal{D}\right\rvert}\sum_{(\mathbf{x}_i, y_i) \in \mathcal{D}} f(\mathbf{w}, \mathbf{x}_i, y_i),
\end{equation}
where $\mathcal{D}$ is the dataset, and $f(\mathbf{w},\mathbf{x}, y)$ is the cross entropy (CE) to measure the sample-wise error over the model, $\mathbf{w}$, w.r.t. sample $\mathbf{x}$ and its true class, $y$ \cite{goodfellow2016deep}. Then, the SGD can be used to minimize the global loss function as follows
\begin{equation}
\mathbf{w}(\ell+1)=\mathbf{w}(\ell)-\eta \frac{\partial F(\mathbf{w})}{\partial \mathbf{w}}\vert_{\mathbf{w}=\mathbf{w}(\ell)}, \label{SGD}
\end{equation}
where $\eta$ is the learning rate and $\mathbf{w}(\ell)$ is the model in training round $\ell$.
Despite their impressive performance in implementing various learning tasks, DNNs have long been regarded as black-boxes \cite{montavon2018methods,rumelhart1986learning}. It is challenging to interpret how the data is transformed as it passes through the DNNs and what the underlying mechanisms are.

FL has been adopted to deal with data privacy concerns associated with training the black-box DNNs at the edge. Instead of uploading the original dataset directly, FL focuses on transmitting model updates to renew the global model through multiple rounds of communication \cite{lim2020federated}.
Specifically, in round $\ell \in \mathcal{L}=\{1, \ 2, \ \dots, \ L\}$, the edge server broadcasts the global model, $\mathbf{w}(\ell)$, to edge devices. Let $F_k(\mathbf{w})$ be the local loss function over a local dataset, $\mathcal{D}_k$, (assuming uniform sizes) at device $k \in \mathcal{K}=\{1, \ 2, \ \dots, \ K\}$. Each device $k$ calculates the gradient of the $F_k(\mathbf{w})$ w.r.t. $\mathbf{w}(\ell)$ based on $\mathcal{D}_k$, and the local model is updated as\footnote{For ease of exposition, here we narrate FedSGD, a special case of FedAvg. It is assumed that in each communication round, there is only one epoch of training for each client model, and the full local dataset is treated as a mini-batch \cite{mcmahan2017communication}.} 
\begin{equation}
\mathbf{w}_k(\ell+1)=\mathbf{w}_k(\ell)-\eta \frac{\partial F_k(\mathbf{w})}{\partial \mathbf{w}}\vert_{\mathbf{w}=\mathbf{w}(\ell)}. \label{gd_cal}
\end{equation}
Subsequently, each edge device uploads the updated model $\mathbf{w}_k(\ell+1)$ to the edge server, and the edge server aggregates the models using the arithmetic mean as
\begin{equation}
\mathbf{w}(\ell+1)=\frac{1}{K}\sum_{k=1}^{K}\mathbf{w}_k(\ell+1). \label{model_ave}
\end{equation}

The procedures of \eqref{gd_cal}-\eqref{model_ave} are iteratively repeated until convergence or the maximal round number $L$ is reached. However, significant communication latency is incurred for two reasons. First, the entire model needs to be transmitted by every device in each communication round. Second, numerous rounds are generally required to achieve convergence, due to the randomness in parameter initialization and SGD.

\subsection{White-box NNs via Forward-only Propagation}\label{white-box}

\begin{figure*}[!t]
\centering
        \subfigure[Interpretation of the parameter impacts and a feature space with $\epsilon$-balls packed.]{\includegraphics[width=0.42\linewidth]{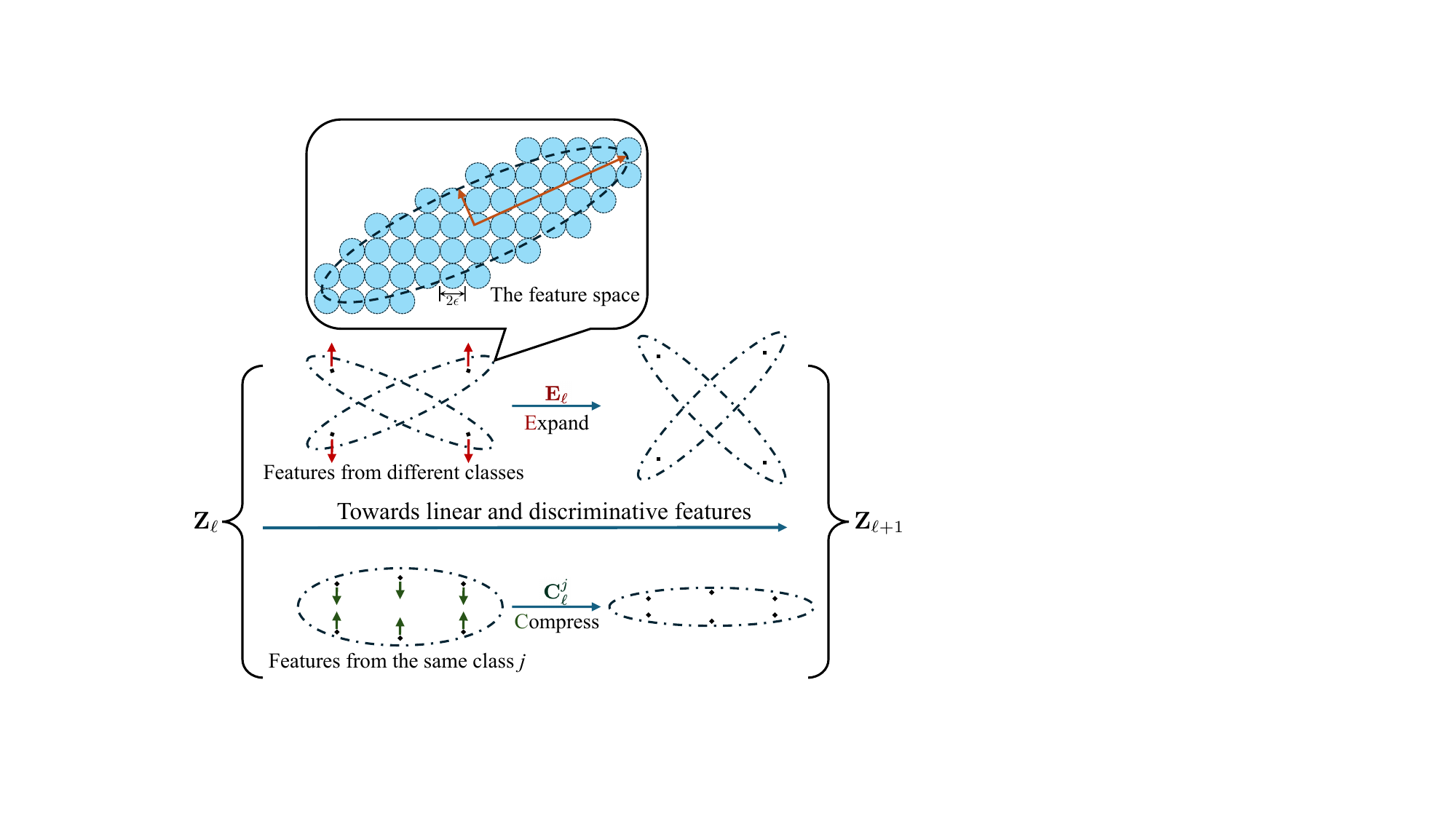}\label{parameter impact}}\hspace{0.1\textwidth}
    \subfigure[The structure of ReduNet's $\ell$-th layer.]{\includegraphics[width=0.3\linewidth]{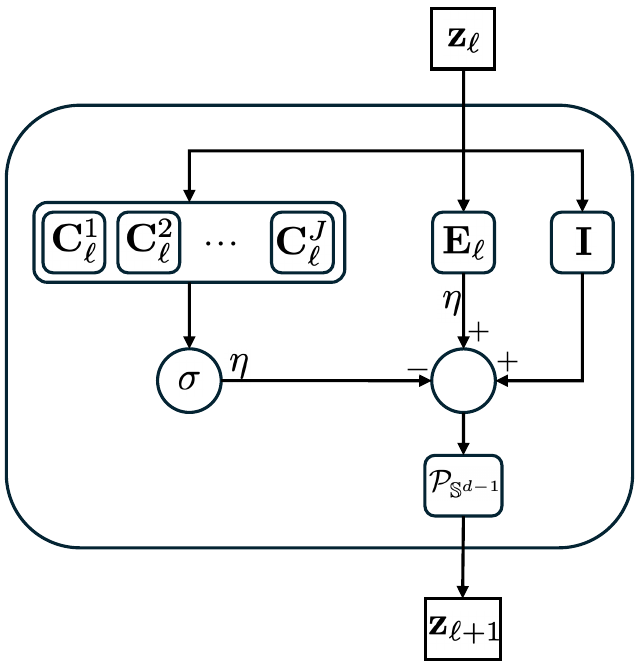}\label{layer structure}}  
    \caption{Illustration of the ReduNet with forward-only propagation.}
    \captionsetup{justification=justified}
    \label{ReduNet}
  \end{figure*}

The training of DNNs has been believed to follow the \emph{parsimony} principle, whose goal is to learn a mapping $\phi(\cdot,\boldsymbol{\theta}_1)$ with parameters $\boldsymbol{\theta}_1$ to transform data $\mathbf{x}$ to a more compact and structured feature $\mathbf{z}$, facilitating downstream tasks \cite{ma2022principles}. Taking classification as an example, 
after the feature $\mathbf{z}$ is obtained, a classifier $\psi(\cdot,\boldsymbol{\theta}_2)$ with parameters $\boldsymbol{\theta}_2$ (see, e.g, \cite{he2016deep}) is then used to predict its class $y$ \cite{chan2022redunet}. The entire pipeline is given as 
$
    \mathbf{x} \overset{\phi(\mathbf{x},\boldsymbol{\theta}_1)}{\xrightarrow{\hspace*{0.8cm}}}\mathbf{z} \overset{\psi(\mathbf{z},\boldsymbol{\theta}_2)}{\xrightarrow{\hspace*{0.8cm}}} y.\label{pipeline}
$
However, the mapping, $\phi(\cdot,\boldsymbol{\theta}_1)$, and the classifier, $\psi(\cdot,\boldsymbol{\theta}_2)$, are jointly optimized in black-box learning, without considering the features' distribution and characteristics.
In contrast, white-box learning aims to find a mapping, $\phi(\cdot,\boldsymbol{\theta}_1)$, that produces $\mathbf{Z}$ with the following linear discriminative properties. Features $\mathbf{Z}$ belonging to different classes exhibit low correlation, indicating that they occupy distinct sub-spaces (ideally orthogonal) and collectively span a large feature space. Conversely, features $\mathbf{Z}^j$ from the same class $j \in \mathcal{J}=\{1, \ 2, \ \dots, \ J\}$ exhibit high correlation and span a small feature sub-space\cite{yu2020learning,chan2022redunet}. However, measuring the feature space with a finite number of feature samples presents the first issue. Additionally, how to find the mapping $\phi(\cdot,\boldsymbol{\theta}_1)$ to transform the data to the features that have the linear discriminative properties, becomes the second issue. Finally, the third issue to be investigated is how to classify an unlabeled sample once its features have been obtained. The details of the solutions to the aforementioned three issues are discussed as follows.

\subsubsection{Measuring the Feature Space with Coding Rate}
The \emph{rate-distortion} was introduced in \cite{cover1999elements} to measure the compactness of a random distribution, defined as the minimal binary bits to encode a random variable up to a specific distortion. Fig.~\ref{parameter impact} illustrates a feature space packed with small balls with diameter $2\epsilon$, where the ball number represents the rate-distortion up to distortion $\epsilon$. 
With unknown distribution and limited samples, computing the rate-distortion is typically intractable. Fortunately, distributions with linear discriminative properties allow closed-form expressions for the total bits to encode the samples \cite{ma2007segmentation}.
With enough samples, the average coding length per sample, a.k.a. the \emph{coding rate}, can approximate the rate-distortion, serving as a natural measure of a feature space's volume.

In particular, given data $\mathbf{X}=\left[ \mathbf{x}^{(1)},\ \mathbf{x}^{(2)},\ \dots,\ \mathbf{x}^{(m)} \right] \in \mathbb{R}^{d\times m}$ with $m$ samples and $d$ dimensions, and their latent features $\mathbf{Z}=\left[ \mathbf{z}^{(1)},\ \mathbf{z}^{(2)},\ \dots,\ \mathbf{z}^{(m)} \right] $ with the same shape, the coding rate of features $\mathbf{Z}$ is
\begin{equation}
    R(\mathbf{Z},\epsilon)\triangleq\frac{1}{2}\log\det\left(\mathbf{I}+\alpha\mathbf{Z}\mathbf{Z}^*\right) 
, \label{Coding rate}
\end{equation}
w.r.t. a certain distortion $\epsilon$, where $(\cdot)^*$ denotes the transpose of a matrix or vector and $\alpha=d/{(m\epsilon^2)}$ \cite{ma2007segmentation,chan2022redunet}. 
Similarly, the coding rate of the union of feature sub-spaces belonging to different classes is given by
\begin{equation}
R_c(\mathbf{Z},\epsilon|\mathbf{\Pi})\triangleq\sum_{j=1}^J\frac{\gamma^j}{2}\log\det\left(\mathbf{I}+\alpha^j\mathbf{Z}\mathbf{\Pi}^j\mathbf{Z}^*\right) 
, \label{Coding rate sub}
\end{equation}
where $\alpha^j=d/{(\textrm{tr}(\mathbf{\Pi}^j)\epsilon^2)}$, $\gamma^j={\textrm{tr}(\mathbf{\Pi}^j)}/{m}$. And $\mathbf{\Pi}\triangleq\{\mathbf{\Pi}^j \in \mathbb{R}^{m\times m}\}_{j=1}^J$ is a set of diagonal membership matrices to characterize the associated classes of data samples. For example, if sample $i$ belongs to class $j$, then $\mathbf{\Pi}^j(i,i)=1$, otherwise, $\mathbf{\Pi}^j(i,i)=0$.

\subsubsection{Constructing ReduNet via MCR$^2$}
With \eqref{Coding rate} and \eqref{Coding rate sub}, the \emph{coding rate reduction} can be defined as
\begin{equation}
\Delta R(\mathbf{Z},\epsilon|\mathbf{\Pi})\triangleq R(\mathbf{Z},\epsilon)-R_c(\mathbf{Z},\epsilon|\mathbf{\Pi})
. \label{Coding rate reduction}
\end{equation}
The linear discriminative properties call for a large volume of the whole feature space, $R(\mathbf{Z},\epsilon)$, and a small volume of the individual feature spaces, $R_c(\mathbf{Z},\epsilon|\mathbf{\Pi})$, which necessities maximizing $\Delta R(\mathbf{Z},\epsilon|\mathbf{\Pi})$ w.r.t. normalized features $\mathbf{Z}$. Meanwhile, a mapping $\phi(\cdot,\boldsymbol{\theta}_1)$ is needed to transform original data $\mathbf{X}$ to features $\mathbf{Z(\boldsymbol{\theta}}_1)=\phi(\mathbf{X},\boldsymbol{\theta}_1)$. This is called maximal coding rate reduction (MCR$^2$), formulated as: $\max_{\boldsymbol{\theta}_1}\Delta R(\mathbf{Z},\epsilon|\mathbf{\Pi}), \ \mathrm{s.t.} \left\lVert \mathbf{Z}^j(\boldsymbol{\theta}_1)\right\rVert_F^2=\textrm{tr}(\mathbf{\Pi}^j). $
The projected gradient ascent scheme \cite{chan2022redunet} works for it:
\begin{equation}
    \mathbf{Z}_{\ell+1} = \mathcal{P}_{\mathbb{S}^{d-1}}(\mathbf{Z}_{\ell}+\tilde{\eta}\frac{\partial \Delta R}{\partial \mathbf{Z}}\vert_{\mathbf{Z}=\mathbf{Z}_{\ell}}), \ \ell=1, \ 2, \ \dots, \ L,\label{PGA}
\end{equation}
where $\tilde{\eta}$ is the learning rate (to be elaborated in the sequel), $L$ is the number of transformations, $\mathbf{Z}_\ell$ are the features after ($\ell-1$) transformations, and $\mathcal{P}_{\mathbb{S}^{d-1}}(\cdot)$ denotes the projection operation which projects vectors to the unit sphere $\mathbb{S}^{d-1}$ for normalization. Specifically,
 $\mathbf{Z}_{1}=\mathcal{P}_{\mathbb{S}^{d-1}} (\mathbf{X}) \in \mathbb{R}^{d\times m}$.

To better understand the gradient, with \eqref{Coding rate}-\eqref{Coding rate reduction}, the gradient in \eqref{PGA} is calculated as
\begin{equation}
    \begin{split}
    &\frac{\partial \Delta R}{\partial \mathbf{Z}}\vert_{\mathbf{Z}=\mathbf{Z}_{\ell}}
    =\alpha(\mathbf{I}+\alpha \mathbf{Z}_{\ell}\mathbf{Z}_{\ell}^*)^{-1}\mathbf{Z}_{\ell}\\
&\quad \quad \quad \quad \quad \quad-\sum_{j=1}^{J}\gamma^j\alpha^j(\mathbf{I}+\alpha^j \mathbf{Z}_{\ell}\mathbf{\Pi}^j\mathbf{Z}_{\ell}^*)^{-1}\mathbf{Z}_{\ell}\mathbf{\Pi}^j.\label{PGA_gradient}
\end{split}
\end{equation}
For simplicity, we denote $\mathbf{E}_{\ell}\triangleq(\mathbf{I}+\alpha \mathbf{Z}_{\ell}\mathbf{Z}_{\ell}^*)^{-1}$ and $\mathbf{C}_{\ell}^j\triangleq(\mathbf{I}+\alpha^j \mathbf{Z}_{\ell}\mathbf{\Pi}^j\mathbf{Z}_{\ell}^*)^{-1}$ as part of the gradient information\footnote{Note that the definitions presented here differ slightly from the original ones in \cite{chan2022redunet}, as the coefficients $\alpha$ and $\alpha_j$ are omitted. This adjustment is due to the fact that $\alpha$ and $\alpha_j$ are related to the number of samples; an increase in the number of samples decreases them and causes the gradients to approach zero, which is an undesirable outcome that requires correction. We remark that this scaling coefficient does not influence the main results, and we safely make this modification.}. Then, \eqref{PGA_gradient} becomes
\begin{equation}
    \frac{\partial \Delta R}{\partial \mathbf{Z}}\vert_{\mathbf{Z}=\mathbf{Z}_{\ell}}=\alpha(\mathbf{E}_{\ell}\mathbf{Z}_{\ell}-\sum_{j=1}^{J}\mathbf{C}_{\ell}^j\mathbf{Z}_{\ell}\mathbf{\Pi}^j).\label{PGA_gradient_2}
\end{equation}
Hence, the increment in \eqref{PGA} becomes $\tilde{\eta}\alpha(\mathbf{E}_{\ell}\mathbf{Z}_{\ell}-\sum_{j=1}^{J}\mathbf{C}_{\ell}^j\mathbf{Z}_{\ell}\mathbf{\Pi}^j)$, and we denote $\eta\triangleq\tilde{\eta}\alpha$, or equivalently $\tilde{\eta}=\eta/\alpha$. Here, $\eta$ is a fixed learning rate, while $\tilde{\eta}$ is a variable learning rate adjusted w.r.t. $\alpha$ (which is related to the number of samples).

As $\mathbf{E}_{\ell}$ and $\mathbf{C}_{\ell}^j$ are from \eqref{Coding rate} and \eqref{Coding rate sub} respectively, $\mathbf{E}_{\ell}$ forces $\mathbf{Z}_{\ell}$ from different classes to diverge while $\mathbf{C}_{\ell}^j$ compresses $\mathbf{Z}_{\ell}^j$ from the same class $j$, and become $\mathbf{Z}_{\ell+1}$, as shown in Fig.~\ref{parameter impact}. Each transformation enhances the features' linear discriminative properties. The transformation with the matrices $\mathbf{E}_{\ell}$ and $\mathbf{C}_{\ell}^j$ can be considered as the effect of one layer of a neural network, called \emph{ReduNet}, whose layer structure is shown in Fig.~\ref{layer structure}. It is constructed via forward-only propagation, where the calculation of $\mathbf{E}_{\ell}$ and $\mathbf{C}_{\ell}^j$ alternates with the feature transformation.
Upon obtaining $\{\mathbf{E}_\ell\}_{\ell=1}^{L}$ and $\{\mathbf{C}_\ell^j\}_{j=1,\ell=1}^{J,L}$, the training is finished.
Since $\mathbf{E}_{\ell}$ and $\mathbf{C}_{\ell}^j$ are derived from the rigorous mathematical principle of MCR$^2$ and their effects are fully interpretable, ReduNet earns the label of \emph{white-box} \cite{chan2022redunet}.

\subsubsection{Employing ReduNet for Inference}
As for inference, considering an unlabeled sample $\mathbf{x}$ and its transformed feature $\mathbf{z}_\ell$ in layer $\ell$, the gradient is ($\mathbf{E}_{\ell}\mathbf{z}_{\ell}-\sum_{j=1}^{J}\gamma^j\mathbf{C}_{\ell}^j\mathbf{z}_{\ell}\boldsymbol{\pi}^j(\mathbf{z}_\ell)$),
where $\boldsymbol{\pi}(\mathbf{z}_\ell)$ is the probability distribution vector of $\mathbf{z}_\ell$.
Building upon the insight from \eqref{PGA_gradient}, gradient $-\mathbf{C}_{\ell}^j\mathbf{z}_{\ell} $ guides $\mathbf{z}_\ell$ towards the sub-space of its true class, which makes $\left\lVert\mathbf{C}_{\ell}^j\mathbf{z}_{\ell} \right\rVert $ small if $\mathbf{z}_\ell$ belongs to class $j$ and large otherwise. This facilitates the estimation of $\boldsymbol{\pi}^j(\mathbf{z}_\ell)$ using softmax as
\begin{align}
\hat{\boldsymbol{\pi}}^j(\mathbf{z}_\ell)\triangleq&\sigma\bigg(\Big[\left\lVert\mathbf{C}_{\ell}^1\mathbf{z}_{\ell} \right\rVert), \ \left\lVert\mathbf{C}_{\ell}^2\mathbf{z}_{\ell} \right\rVert), \ \dots, \ \left\lVert\mathbf{C}_{\ell}^J\mathbf{z}_{\ell} \right\rVert)\Big]\bigg)^j\\
=&{\mathrm{exp}\Big(-\lambda\left\lVert\mathbf{C}_{\ell}^j\mathbf{z}_{\ell} \right\rVert\Big)}/{\sum_{j=1}^J\mathrm{exp}\Big(-\lambda\left\lVert\mathbf{C}_{\ell}^j\mathbf{z}_{\ell} \right\rVert\Big)},\label{Exp of pk}
\end{align}
where $\lambda$ is a hyperparameter. Then a classifier for ReduNet is given by $\hat{j}=\mathrm{arg}\max_{j\in \mathcal{J}}(\hat{\boldsymbol{\pi}}(\mathbf{z}_L))$.

As ReduNet is constructed layer by layer, a novel FL framework can be designed to enable layer-wise construction and update, as presented in the following sections.

\section{System Model}\label{system model}

Consider a general FL system in which $K$ edge devices with their local datasets aim to learn the optimal NN parameters as coordinated by the edge server over a total of $L$ communication rounds. 
Similar to the traditional FL procedure, in each communication round, local models are updated based on local datasets, the updated parameters are uploaded to the edge server for aggregation, and the aggregated parameters are subsequently broadcast to edge devices for updating local models (elaborated in Section \ref{sec LoLaFL framework}). The transmission process in each communication round is described as follows.

The orthogonal frequency-division multiple access (OFDMA) is adopted, where the available bandwidth $B$ is divided into $M$ orthogonal subchannels, and each edge device is assigned $M/K$ sub-channels to avoid the interference\cite{goldsmith2005wireless,zhu2019broadband}. At edge device $k$, local parameters $\mathbf{g}_k \in \mathbb{R}^q$ are to be uploaded. Each parameter is quantized into Q bits by uniform quantization as in\cite{liu2023over} which are then modulated into symbols. The $i$-th symbol received at the server is given by
\begin{equation} y_{i,k}=h_{k}\sqrt{p_{k}}x_{i,k}+n_{i,k},\label{x_transmission_digital}
\end{equation}
where
$x_{i,k}$ is the $i$-th symbol from edge device $k$, $h_k$ is the channel coefficient between device $k$ and server, $p_k$ is the associated power control policy, and $n_{i,k}\sim \mathcal{CN}(0,\nu_n^2)$ is the independent and identically distributed (IID) additive white Gaussian noise (AWGN). We assume a slow fading channel where $h_k$ remains constant over a single uploading round and is assumed to be known to both sides. We model $h_k$ as Rayleigh fading with $h_k\sim \mathcal{CN}(0,1)$, where the coefficients are IID across different devices and different communication rounds \cite{liu2023over,wang2024spectrum}.

In FL, model aggregation is implemented after all devices have completed uploading their local models. Consequently, poor channel conditions can impede the local model uploading process on some devices, thereby increasing overall latency.
To mitigate fading, we adopt the truncated power control policy, as in \cite{zhu2019broadband}:
\begin{equation}
  p_k =
    \begin{cases}
      \rho_0 /|h_k|^2, & |h_k|^2\geq\tau,\\
      0, & |h_k|^2<\tau,\label{power control}
    \end{cases}       
\end{equation}
where $\rho_0$ is a scaling factor to meet the power constraint in the sequel, and $\tau$ is the power cut-off threshold to avoid deep fading.
The power constraint for each subchannel is $\mathbb{E}[p_k]\leq \frac{KP_0}{M}$, with $P_0$ being the power budget per device.
Since $h_k\sim \mathcal{CN}(0,1)$, $|h_k|^2$ follows an exponential distribution with unit mean. Therefore, analyzing the expectation of $p_k$, we have
\begin{equation}
\begin{split}
     \mathbb{E}[p_k]=&\int_\tau^\infty\frac{\rho_0}{x}\exp(-x)\,dx\\ =&\rho_0\int_\tau^\infty\frac{1}{x}\exp(-x)\,dx.\label{exp of pk}
\end{split}
\end{equation}
Hence, we can derive the exact value of $\rho_0=KP_0/(M\mathrm{Ei}(\tau))$, with $\mathrm{Ei}(x)=\int_{x}^{\infty}\frac{1}{s}\mathrm{exp}(-s) \,ds$ \cite{abramowitz1968handbook}. This policy can result in an outage probability of $\xi \triangleq \mathrm{Pr}(|h_k|^2<\tau)=1-\mathrm{exp}(-\tau)$.
According to the above settings, 
when $|h_k|^2\geq\tau$, the receive SNR is given by $\frac{|h_k|^2p_k}{\nu_n^2}=\frac{\rho_0}{\nu_n^2}=\frac{KP_0}{M \nu_n^2\mathrm{Ei}(\tau)}$. Therefore, the transmission rate of device $k$ is given by
 \begin{equation}
     r_k=\frac{B}{K}\log_2\Big(1+\frac{KP_0}{M \nu_n^2 \mathrm{Ei}(\tau)}\Big).
 \end{equation}
Then, the uploading communication latency (in seconds) for device $k$ in round $\ell$ is given by
\begin{equation}
T_{\mathrm{comm},\ell,k}=\frac{KqQ}{B\log_2\Big(1+\frac{KP_0}{M \nu_n^2 \mathrm{Ei}(\tau)}\Big)}.\label{latency}
\end{equation}
For devices whose channels fail to meet the threshold $\tau$, they give up transmission without repeated attempts. Thus, no additional retransmission latency is incurred. The resultant loss could degrade the learning performance, which will be investigated in experiments.

The edge server demodulates the received symbols to recover the bit streams and reconstruct the local parameters $\bar{\mathbf{g}}_k$ for calculating the global ones $\bar{\mathbf{g}}$. Subsequently, the global parameters are broadcast to devices to replace their local models. Since the edge server typically has higher transmit power and full downlink bandwidth availability, the broadcasting latency is negligible compared to that of uploading and is thus omitted from our analysis \cite{yang2020energy}. 

\section{LoLaFL via Forward-only Propagation} \label{LoLaFL}

In this section, we propose a novel FL framework for achieving low-latency edge learning based on the white-box NN introduced in Section \ref{white-box}. 
First, the model uploading and aggregation processes of the proposed framework based on the forward-only propagation algorithm are introduced. Then, two novel nonlinear aggregation methods are presented.

\begin{figure*}[!t]
\centering
    \subfigure[The model updating and aggregation processes.]{\includegraphics[width=0.67\linewidth]{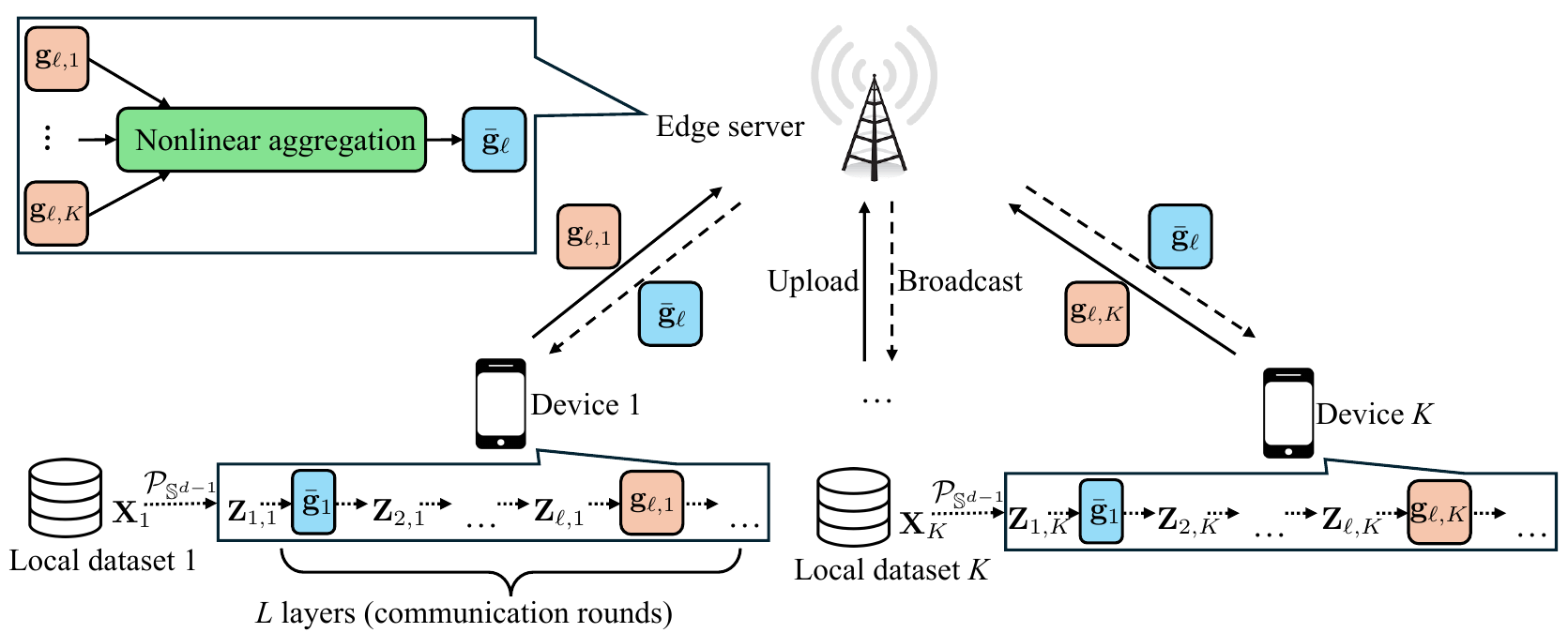}\label{FL system}}
    \subfigure[Local training details in round $\ell$ at device $k$.]{\includegraphics[width=0.32\linewidth]{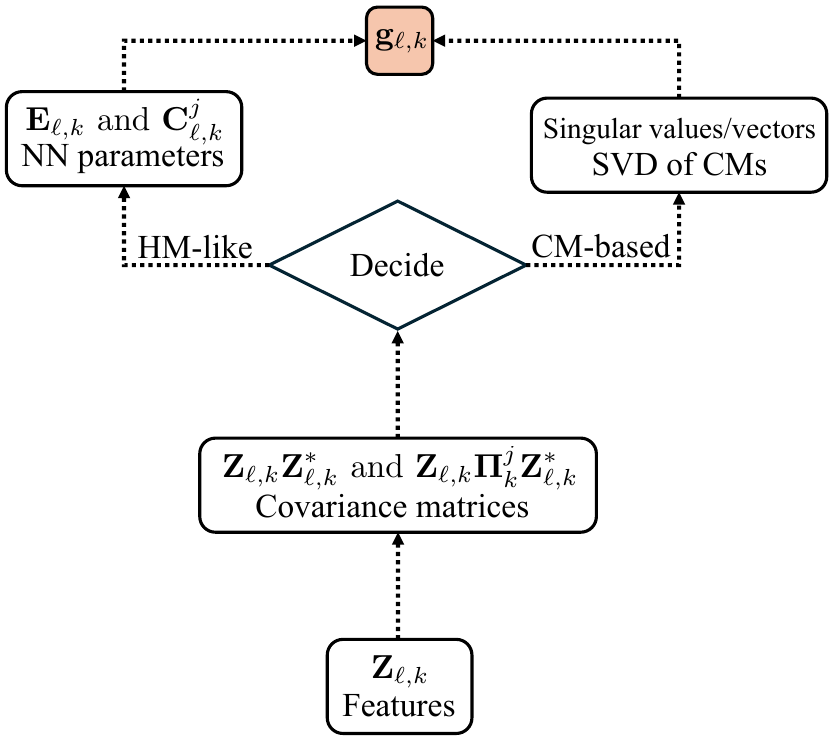}\label{local training}}  
    \caption{The LoLaFL framework.}
    \captionsetup{justification=justified}
    \label{LoLaFL framework}
  \end{figure*}

\subsection{The LoLaFL Framework}\label{sec LoLaFL framework}

We propose a novel LoLaFL framework as shown in 
Fig.~\ref{FL system}. Unlike traditional FL where the whole model is exchanged between the edge devices and the server, LoLaFL enables the white-box NNs to be constructed and updated in a layer-wise manner. The details are provided as follows.

\subsubsection{Layer-wise Construction}

This part corresponds to the local training in traditional FL, but the approach is fundamentally different.
In each communication round in LoLaFL, the parameters of a single layer are calculated directly based on the latest features at each device.
Initially, the local data samples at device $k$ are normalized, outputting the features, i.e., $\mathbf{Z}_{1,k}=\mathcal{P}_{\mathbb{S}^{d-1}} (\mathbf{X}_k) \in \mathbb{R}^{d\times m_k}$, where $m_k$ is the number of samples at device $k$, and their associated classes are characterized by the diagonal membership matrices, $\{\mathbf{\Pi}_k^j \in \mathbb{R}^{m_k\times m_k}\}_{j=1}^J$, as defined in \eqref{Coding rate sub}.
In communication round $\ell\in \mathcal{L}=\{1, \ 2, \ \dots, \ L\}$, the local feature samples, $\mathbf{Z}_{\ell,k}$, and their corresponding membership matrices, $\{\mathbf{\Pi}_k^j\}_{j=1}^J$, at edge device $k$, are utilized to calculate the NN parameters of ReduNet's $\ell$-th layer according to \eqref{PGA_gradient}. In other words,
\begin{equation}
    \mathbf{E}_{\ell,k}\triangleq (\mathbf{I}+\alpha_k \mathbf{Z}_{\ell,k}\mathbf{Z}_{\ell,k}^{*})^{-1},\label{El_k}
\end{equation}
\begin{equation} \mathbf{C}_{\ell,k}^j\triangleq(\mathbf{I}+\alpha_k^j\mathbf{Z}_{\ell,k}\mathbf{\Pi}_k^j\mathbf{Z}_{\ell,k}^{*})^{-1}.\label{Cl_jk}
\end{equation}
In the formulae, $\alpha_k=d/{(m_k\epsilon^2)}$, $\alpha_k^j=d/{(\textrm{tr}(\mathbf{\Pi}_k^j)\epsilon^2)}$, and $\gamma_k^j={\textrm{tr}(\mathbf{\Pi}_k^j)}/{m_k}$ are the local coefficients.
We assume that all edge devices and the edge server share the information of $m$ and $\textrm{tr}(\mathbf{\Pi}^j)$, and have an identical setting of $\epsilon$, which means all edge devices and the edge server can calculate the global coefficients $\alpha$, $\alpha^j$, and $\gamma^j$ individually. Additionally, the edge server is aware of the $m_k$ and $\textrm{tr}(\mathbf{\Pi}_k^j)$. The local training process of LoLaFL is shown in Fig.~\ref{local training}.

\subsubsection{Layer-wise Transmission and Aggregation}

Different from traditional FL that focuses on the whole model, LoLaFL only uploads and aggregates one model layer per communication round. Specifically, after the $\ell$-th NN layer with parameters, $\mathbf{g}_{\ell,k}$, is constructed, device $k$ aims to transmit this layer's parameters to the server for aggregation. Depending on the special white-box structures given in \eqref{El_k} and \eqref{Cl_jk}, the transmitted parameters can be either the exact NN parameters or the latent covariance matrices (CMs) of features. This calls for different aggregation designs, which will be introduced in the following subsections in detail. Here, we let the transmitted parameters be $\mathbf{g}_{\ell,k}$ to illustrate the LoLaFL framework as shown in Fig.~\ref{FL system}. When the server receives $\{\mathbf{g}_{\ell,k}\}_{k=1}^K$, the global parameters $\Bar{\mathbf{g}}_{\ell}$ are calculated and updated, which are then broadcast to edge devices. At edge device $k$, its local parameters, $\mathbf{g}_{\ell,k}$, are replaced by the received global parameters $\Bar{\mathbf{g}}_{\ell}$. Afterwards, the local features $\mathbf{Z}_{\ell,k}$ are input into the $\ell$-th layer with parameters, $\Bar{\mathbf{g}}_{\ell}$, to output $\mathbf{Z}_{\ell+1,k}$ for constructing the ($\ell+1$)-th layer.
The LoLaFL algorithm is summarized in Algorithm \ref{FRedu Training}.
\begin{algorithm}
 \caption{Proposed LoLaFL Algorithm}
 \label{FRedu Training}
 \begin{algorithmic}[1]
 \renewcommand{\algorithmicrequire}{\textbf{Input:}}
 \renewcommand{\algorithmicensure}{\textbf{Output:}}
 \REQUIRE $\{\mathbf{X}_k\in \mathbb{R}^{d\times m_k}\}_{k=1}^K$, $\{\mathbf{\Pi}_{k}^j\in \mathbb{R} ^{m_k\times m_k}\}_{j=1,k=1}^{J,K}$, $\epsilon$, $\lambda$, learning rate $\eta$, layer number $L$, channel inversion threshold $\tau$ (, SVD threshold $\beta_0$).
\textit{Initialization}: $\{\mathbf{Z}_{1,k}=\mathcal{P}_{\mathbb{S}^{d-1}} (\mathbf{X}_k)\in \mathbb{R}^{d\times m_k}\}_{k=1}^K$, $\alpha=d/{(m\epsilon^2)}$, $\{\alpha^j=d/{(\textrm{tr}(\mathbf{\Pi}^j)\epsilon^2)}\}_{j=1}^{J}$, $\{\gamma^j={\textrm{tr}(\mathbf{\Pi}^j)}/{m}\}_{j=1}^{J}$.
  \FOR {$\ell = 1$ to $L$}
        \FOR {$k = 1$ to $K$}
                \STATE Local NN parameter calculation (\eqref{El_k} and \eqref{Cl_jk}, for the HM-like scheme) or local SVD of covariance matrices calculation (\eqref{ZZ^k approx} and \eqref{ZPZ^k approx}, for the CM-based scheme).
            \IF{Deep fading ($|h_k|^2<\tau$)}
            \STATE Device $k$ quits parameters uploading in this round.
            \ELSE
            \STATE Local NN parameters (for the HM-like scheme) or decomposed covariance matrices (for the CM-based scheme) uploading.
            \ENDIF
            \ENDFOR
        \STATE Aggregation with local NN parameters (\eqref{El} and \eqref{Cl_j}, for the HM-like scheme), or with local covariance matrices (\eqref{ZZ^k reconstruct} and \eqref{R^j global approx}, for the CM-based scheme), for global ones. 
        \STATE (Global NN parameter calculation \eqref{PGA_gradient} for the CM-based scheme, if needed.)
        \STATE Global NN parameters or decomposed covariance matrices broadcasting.
        \FOR {$k = 1$ to $K$}
            \STATE (NN parameter calculation \eqref{El_k}, \eqref{Cl_jk} for the CM-based scheme.)
            \STATE Feature transformation \eqref{PGA}.
        \ENDFOR
  \ENDFOR
\ENSURE Learned parameters of $\{\mathbf{E}_\ell\}_{\ell=1}^{L}$ and  $\{\mathbf{C}_\ell^j\}_{j=1,\ell=1}^{J,L}$.
 \end{algorithmic} 
 \end{algorithm}

\subsection{Harmonic-mean-Like Aggregation}\label{HM}

In this subsection, the parameters to be exchanged between edge devices and the edge server are the parameters of the white-box NN, i.e., $\mathbf{g}_{\ell,k}=\{\mathbf{E}_{\ell,k}\}\cup \{\mathbf{C}_{\ell,k}^j\}_{j=1}^J$. However, the FedAvg in traditional FL is not optimal for the aggregation in this scenario, because the NN parameters are derived from the features with nonlinear mappings. Hence, a novel compatible aggregation scheme is designed for LoLaFL as follows.

Referring to \eqref{El_k} and \eqref{Cl_jk}, the local NN parameters, $\{\mathbf{E}_{\ell,k}\}\cup \{\mathbf{C}_{\ell,k}^j\}_{j=1}^J$, are determined by the CMs of local features, i.e., $\mathbf{R}_{\ell,k}\triangleq\mathbf{Z}_{\ell,k}\mathbf{Z}_{\ell,k}^{*}$ and $\mathbf{R}_{\ell,k}^j\triangleq\mathbf{Z}_{\ell,k}\mathbf{\Pi}_{k}^j\mathbf{Z}_{\ell,k}^{*}$. And referring to \eqref{PGA_gradient}, the global NN parameters, $\{\mathbf{E}_{\ell}\}\cup \{\mathbf{C}_{\ell}^j\}_{j=1}^J$, are determined by the CMs of global features, i.e., $\Bar{\mathbf{R}}_{\ell}\triangleq\mathbf{Z}_{\ell}\mathbf{Z}_{\ell}^*$ and $\Bar{\mathbf{R}}^j_{\ell}\triangleq\mathbf{Z}_{\ell}\mathbf{\Pi}^j\mathbf{Z}_{\ell}^*$. Therefore, aggregation of $\{\mathbf{E}_{\ell,k}\}\cup \{\mathbf{C}_{\ell,k}^j\}_{j=1}^J$ fundamentally requires aggregation of the CMs of local features.
To this end, we obtain the following results.

\textit{\textbf{Lemma 1}} 
In each communication round $\ell$, the CMs of global features can be decomposed as the summation of the CMs of local features. In other words,
\begin{equation}
\Bar{\mathbf{R}}_{\ell}=\sum_{k=1}^K\mathbf{R}_{\ell,k}
\qquad \mathrm{and} \qquad \Bar{\mathbf{R}}^j_{\ell}=\sum_{k=1}^K\mathbf{R}^j_{\ell,k}.\label{lemma1,ZPZ}
\end{equation}

\textit{Proof:} See Appendix A.

While technical proof is available in the appendix, we offer insights into why this holds true. Taking the first formula in \eqref{lemma1,ZPZ} as an example, the elements of $\Bar{\mathbf{R}}_{\ell}$ represent the energy. Specifically, its diagonal elements represent the energy of each feature dimension, while the off-diagonal elements represent shared energy (correlation) between feature dimensions. The calculation of energy is based on selected feature samples, which inherently makes it decomposable. This enables potential parallel computation with multiple edge devices for the CMs of global features, which will be discussed later.

\textit{\textbf{Proposition 1} (HM-like aggregation):} In each communication round $\ell$, the global NN parameters, $\Bar{\mathbf{g}}_{\ell}=\{\Bar{\mathbf{E}}_{\ell}\}\cup \{\Bar{\mathbf{C}}_{\ell}^j\}_{j=1}^J$, can be calculated directly with local NN parameters, $\{\mathbf{g}_{\ell,k}\}_{k=1}^K$, as
\begin{equation}
\Bar{\mathbf{E}}_{\ell}=\left(\sum_{k = 1}^{K} \omega_k(\mathbf{E}_{\ell,k})^{-1}\right)^{-1},\label{El}
\end{equation}
\begin{equation}
\Bar{\mathbf{C}}_{\ell}^{j}=\left(\sum_{k = 1}^{K} \omega_k^j(\mathbf{C}_{\ell,k}^j)^{-1}\right)^{-1},\label{Cl_j}
\end{equation}
where $\omega_k\triangleq{m_k}/{m}$ and $\omega_k^j\triangleq{\textrm{tr}(\mathbf{\Pi}_k^j)}/{\textrm{tr}(\mathbf{\Pi}_k})$. We have $\sum_{k = 1}^{K} \omega_k=1$
and $\sum_{k = 1}^{K} \omega_k^j=1$ for any $j\in\mathcal{J}$. 

\textit{Proof:} See Appendix B.

The preceding results demonstrate how to calculate the global NN parameters from the local NN parameters in LoLaFL. Specifically, if we treat the matrices as numbers and the matrix inversion as reciprocal, these two formulae suggest that the global NN parameters are like the weighted \emph{harmonic mean} of the corresponding local NN parameters, with $\omega_k$ and $\omega_j$ being the weights\footnote{We acknowledge that considering the harmonic mean as an aggregation method in traditional FL could also be a promising direction, wherein model updates are inverted, averaged, and then inverted again, all in an element-wise manner. The advantages of this aggregation method are twofold: 1) the harmonic mean is not sensitive to large elements; and 2) the harmonic mean is no larger than the arithmetic mean. These two factors both diminish the likelihood of exploding gradients. However, the numerical stability, measures to address potential instability, and convergence analysis require further investigation.}. This nonlinear aggregation inherently results from the fact that the NN parameters are calculated from features with nonlinear transformations. Note that when data are uniformly distributed across devices, \eqref{El} further reduces to $\Bar{\mathbf{E}}_{\ell}=\left(\frac{1}{K}\sum_{k = 1}^{K} (\mathbf{E}_{\ell,k})^{-1}\right)^{-1}$. And when data belonging to classes $j$ are uniformly distributed across devices, \eqref{Cl_j} further reduces to $\Bar{\mathbf{C}}_{\ell}^{j}=\left(\frac{1}{K}\sum_{k = 1}^{K} (\mathbf{C}_{\ell,k}^j)^{-1}\right)^{-1}$. They are the standard forms of the harmonic mean.

In the $\ell$-th communication round, the procedures are discussed as follows.
Firstly, the local NN parameters, $\mathbf{E}_{\ell,k}$ and $\mathbf{C}_{\ell,k}^j$, are calculated with local features using \eqref{El_k} and \eqref{Cl_jk}.
Then, the local NN parameters at each edge device are uploaded, and the edge server receives the local NN parameters as ($\Bar{\mathbf{E}}_{\ell,k}=\mathbf{E}_{\ell,k}+\mathbf{N}_{\ell,k}$) and ($\Bar{\mathbf{C}}_{\ell,k}^j=\mathbf{C}_{\ell,k}^j+\mathbf{N}_{\ell,k}^j$), with the distortions, $\mathbf{N}_{\ell,k}$ and $\mathbf{N}_{\ell,k}^j$, specified in the system model.
After uploading, the global NN parameters, $\Bar{\mathbf{E}}_{\ell}$ and $\Bar{\mathbf{C}}_{\ell}^{j}$, are calculated based on the received local NN parameters, using \eqref{El} and \eqref{Cl_j} by replacing ${\mathbf{E}}_{\ell,k}$ with $\Bar{\mathbf{E}}_{\ell,k}$ and ${\mathbf{C}}_{\ell,k}^j$ with $\Bar{\mathbf{C}}_{\ell,k}^j$.
Subsequently, the global NN parameters $\Bar{\mathbf{E}}_{\ell}$ and $\Bar{\mathbf{C}}_{\ell}^{j}$ are broadcast to all devices.
Finally, each edge device updates its current layer, i.e., setting its current NN parameters as $\mathbf{E}_{\ell,k}={\Bar{\mathbf{E}}}_{\ell}$ and $\mathbf{C}_{\ell,k}^j={\Bar{\mathbf{C}}}_{\ell}^{j}$. They use the new NN parameters to transform the local features using \eqref{PGA}, which prepares for updating the next layer in the following communication round.

\subsection{Covariance-matrix-Based Aggregation}\label{CM}

In this subsection, the parameters to be exchanged between edge devices and the edge server are the collection of the low-rank versions of local CMs, i.e., $\mathbf{g}_{\ell,k}=\{\tilde{\mathbf{R}}_{\ell,k}\}\cup \{\tilde{\mathbf{R}}_{\ell,k}^j\}_{j=1}^J$, the details of which are given in the sequel. We propose this approach because the NN parameters in the HM-like scheme have very high dimensionality and may be difficult to compress. In contrast, these CMs have low-rank structures, resulting from the low-rank structures of the features. This is because ReduNet is making features sparse, so the intrinsic dimensionality of the features is small, as shown in Fig.~\ref{parameter impact}.
Therefore, these CMs can be further compressed, which motivates the design of CM-based aggregation as follows\footnote{We acknowledge the possible adoption of other compression techniques, e.g., sparsification\cite{sattler2019robust} and quantization with fewer bits\cite{zhu2020one}, for further reducing communication latency. However, since they are applicable to both LoLaFL and traditional FL, we choose not to consider them in this paper. Their effects can be explored in the future work.}.
For ease of notation, in the following exposition, the index $\ell$ is omitted whenever no confusion arises.

The procedure of each communication round is described as follows.
Firstly, the local CMs $\mathbf{R}_k$ and $\mathbf{R}_{k}^j$ at each edge device are calculated.
Then, the local CMs at each edge device are decomposed with \emph{singular value decomposition} (SVD) \cite{moon2000mathematical} and approximated to some degree as follows:
\begin{subequations}
    \begin{align}
        \mathbf{R}_{k}\approx \tilde{\mathbf{R}}_{k}=\sum_{i=1}^{s_{k}} \sigma_{i,k} \mathbf{u}_{i,k} \mathbf{v}_{i,k}^{*},\label{ZZ^k approx}\\
        \mathbf{R}_{k}^j\approx \tilde{\mathbf{R}}_{k}^j=\sum_{i=1}^{s_{k}^{j}} \sigma_{i,k}^j \mathbf{u}_{i,k}^j \mathbf{v}_{i,k}^{j*}.\label{ZPZ^k approx}
    \end{align}
\end{subequations}
In the preceding formulae, $s_{k}$ and $s_k^{j}$ are the minimal possible $s$ to remain desired information: $\beta\triangleq{\sum_{i=1}^s\sigma_i}/{\sum_{i=1}^d\sigma_i}\geq \beta_0$, where $\beta$ is the information remaining rate and $\beta_0$ is the threshold.
We define the compression rate $\delta$ as the expected ratio of the number of chosen singular values to the total number of singular values.
Then the singular values and vectors are uploaded as $\Bar{\sigma}_{i,k}=\sigma_{i,k}+n_{i,k}$, $\Bar{\mathbf{u}}_{i,k}=\mathbf{u}_{i,k}+\mathbf{n}_{\mathbf{u},i,k}$, $\Bar{\mathbf{v}}_{i,k}=\mathbf{v}_{i,k}+\mathbf{n}_{\mathbf{v},i,k}$, $\Bar{\sigma}_{i,k}^j=\sigma_{i,k}^j+n_{i,k}^j$, $\Bar{\mathbf{u}}_{i,k}^{j}=\mathbf{u}_{i,k}^{j}+\mathbf{n}_{\mathbf{u},i,k}^{j}$, and $\Bar{\mathbf{v}}_{i,k}^{j}=\mathbf{v}_{i,k}^{j}+\mathbf{n}_{\mathbf{v},i,k}^{j}$, where the distortions are specified in the system model.
Thus the low-rank-approximated CMs can be reconstructed at the edge server as
\begin{subequations}
    \begin{align}
        \Bar{\mathbf{R}}_k=\sum_{i=1}^{s_{k}} \Bar{\sigma}_{i,k} \Bar{\mathbf{u}}_{i,k} \Bar{\mathbf{v}}_{i,k}^{*}\label{ZZ^k reconstruct},\\
        \Bar{\mathbf{R}}_{k}^j=\sum_{i=1}^{s_k^{j}} \Bar{\sigma}_{i,k}^j \Bar{\mathbf{u}}_{i,k}^j \Bar{\mathbf{v}}_{i,k}^{{j}*}.\label{ZPZ^k reconstruct}
    \end{align}
\end{subequations}
Then we can calculate the CMs of global features, $\Bar{\mathbf{R}}$ and $\Bar{\mathbf{R}}^j$, using \eqref{lemma1,ZPZ} by replacing $\mathbf{R}_{\ell,k}$ with $\Bar{\mathbf{R}}_{\ell,k}$ and $\mathbf{R}^j_{\ell,k}$ with $\Bar{\mathbf{R}}^j_{\ell,k}$.
If needed (when the edge server also needs the entire model), the global NN parameters can be calculated using \eqref{PGA_gradient} by replacing $\mathbf{Z}_{\ell}\mathbf{Z}_{\ell}^*$ with $\Bar{\mathbf{R}}$ and $\mathbf{Z}_{\ell}\mathbf{\Pi}^j\mathbf{Z}_{\ell}^*$ with $\Bar{\mathbf{R}}^{j}$.
Again, we can apply low-rank approximation to the global CMs as
\begin{subequations}
    \begin{align}
        \Bar{\mathbf{R}}\approx \tilde{{\mathbf{R}}}= \sum_{i=1}^{{s_0}} {\sigma}_i {{\mathbf{u}}}_i {\mathbf{v}}_i^{*},\label{R global approx}\\
        \Bar{\mathbf{R}}^{j}\approx \tilde{{\mathbf{R}}}^{j}= \sum_{i=1}^{{s}_0^{j}} {\sigma}_i^{j} {\mathbf{u}}_i^{{j}} {\mathbf{v}}_i^{j*}.\label{R^j global approx}
    \end{align}
\end{subequations}
Subsequently, the singular values and singular vectors are broadcast to each edge device. The low-rank-approximated global CMs can be reconstructed at each edge device using \eqref{R global approx} and \eqref{R^j global approx}.
Finally, each edge device calculates the NN parameters using the definition provided in \eqref{PGA_gradient} by replacing $\mathbf{Z}_{\ell}\mathbf{Z}_{\ell}^{*}$ with $\tilde{{\mathbf{R}}}$ and $\mathbf{Z}_{\ell}\mathbf{\Pi}^j\mathbf{Z}_{\ell}^{*}$ with $\tilde{{\mathbf{R}}}^{j}$. The parameters are then utilized to transform the features according to \eqref{PGA}, which prepares for updating the next layer in the following communication round.

\section{Performance Analysis}\label{Theoretic Analysis}

In this section, we first analyze the communication latency and computational complexity of the LoLaFL with a comparison with traditional FL. Next, we provide a proof of the privacy guarantee in LoLaFL.

\subsection{Latency Analysis}\label{latency analysis}

For brevity, we only consider the number of parameters uploaded from each device $k$, from which the communication latency can be easily obtained.
For LoLaFL with HM-like aggregation, in each round, uploading of local parameters yields $(J+1)d^2$. So, the total number of parameters transmitted over $L$ rounds is $L(J+1)d^2$.
For LoLaFL with CM-based aggregation, since SVD is used to reduce the latency, in each round, the uploading of compressed CMs yields $(J+1)(2\delta d^2+\delta d)$. Thus, the total number of parameters transmitted over $L$ rounds is $L(J+1)(2\delta d^2+\delta d)$.
For traditional FL, let $W$ denote the parameter number of the utilized DNN model. In each round, uploading the local parameters yields $W$. And the total number of parameters transmitted over $L$ rounds is $LW$.

As summarized in Table \ref{4schems_comp}, considering the number of parameters to be transmitted and focusing on the dominant part (i.e., terms with $d^2$) in the expressions, the CM-based scheme outperforms the HM-like scheme, as long as $\delta<1/2$. The latency of LoLaFL is proportional to $d^2$ and $J$ while that of traditional FL does not depend on $d$ and $J$. This means that for datasets with high dimensionality and a large number of classes, LoLaFL may not outperform traditional FL.

\renewcommand{\arraystretch}{1.25}
\begin{table*}[t]
\normalsize
\caption{The Summary of Communication Latency (in parameter) and Computational Complexity}
\begin{center}
\resizebox{1.6\columnwidth}{!}{\begin{tabular}{|c|ccc|}
\hline
\multirow{2}{*}{\textbf{Metrics}} & \multicolumn{3}{c|}{\textbf{Comparison of Different Schemes}}                                                                                                                                                  \\ \cline{2-4} 
                                  & \multicolumn{1}{c|}{\textbf{LoLaFL (HM-like)}}                                     & \multicolumn{1}{c|}{\textbf{LoLaFL (CM-based)}}                                                & \textbf{Traditional FL}  \\ \hline
Latency (per device)              & \multicolumn{1}{c|}{$L(J+1)d^2$}                                                   & \multicolumn{1}{c|}{$L(J+1)(2\delta d^2+\delta d)$}                                            & $LW$                     \\ \hline
Complexity (per round)            & \multicolumn{1}{c|}{$\begin{aligned}&O((J+1)(2K+1)d^3\\&+(J+3)md^2)\end{aligned}$} & \multicolumn{1}{c|}{$\begin{aligned}&O((J+1)(2K+1)d^3\\&+[4\delta K+(J+3)m]d^2)\end{aligned}$} & $O(2m((N-1)n^2+(J+d)n))$ \\ \hline
\end{tabular}}
\label{4schems_comp}
\end{center}
\end{table*}

\subsection{Complexity Analysis}\label{complexity}

For computational complexity, we only consider matrix multiplication, matrix inversion, and SVD (if any), as these operations dominate the complexity. Generally, the multiplication of two matrices with shapes ($m\times n$) and ($n\times k$) takes $mnk$ operations. For an invertible $n\times n$ matrix, the computational complexity of calculating its inversion is $O(n^3)$. For an $m\times n$ matrix, the computational complexity of calculating its SVD is $O(mn\min(m,n))$ \cite{golub2013matrix}.

For LoLaFL with HM-like aggregation, in each communication round, according to \eqref{El_k} and \eqref{Cl_jk}, the parameter calculation at edge devices requires $\sum_{k=1}^{K}O(2m_kd^2+(J+1)d^3)=O((J+1)Kd^3+2md^2)$. Based on \eqref{El} and \eqref{Cl_j}, the aggregation at edge server requires $O((J+1)(K+1)d^3)$. According to \eqref{PGA}, feature transformation requires $\sum_{k=1}^{K}O((J+1)m_kd^2)=O((J+1)md^2)$. Combining these operations yields a computational complexity of $O((J+1)(2K+1)d^3+(J+3)md^2)$.

For LoLaFL with CM-based aggregation, in each communication round, the local CM calculation at edge devices requires $\sum_{k=1}^{K}O(2m_kd^2)=O(2md^2)$. According to \eqref{ZZ^k approx} and \eqref{ZPZ^k approx}, the SVD for the local CMs requires $\sum_{k=1}^{K}O((J+1)d^3)=O((J+1)Kd^3)$. The reconstruction process at the edge server requires $\sum_{k=1}^{K}O(2\delta d^2)=O(2\delta Kd^2)$. The aggregation at the edge server can be omitted because only addition is used. According to \eqref{R global approx} and \eqref{R^j global approx}, the SVD for the global CMs requires $O((J+1)d^3)$, and the reconstruction process at the edge devices requires $\sum_{k=1}^{K}O(2\delta d^2)=O(2\delta Kd^2)$. The parameter calculation and feature transformation require $O((J+1)Kd^3)$ and $\sum_{k=1}^{K}O((J+1)m_kd^2)=O((J+1)md^2)$ respectively. Combining these operations yields $O((J+1)(2K+1)d^3+[4\delta K+(J+3)m]d^2)$.

For traditional FL, we analyze a fully-connected NN with $N$ layers, each containing $n$ nodes. During forward propagation, passing $m_k$ samples from the input layer to the first hidden layer incurs $O(m_kdn)$. Passing them through the subsequent ($N-1$) hidden layers yields $O((N-1)m_kn^2)$, and passing them from the last hidden layers to the output layer yields $O(m_kJn)$. The low complexity associated with adding the bias term and calculating the activation function is omitted. Combining these components results in the complexity of forward propagation for device $k$ as $O(m_k(dn+(N-1)n^2+Jn))$, which is equivalent to that of the backpropagation. Therefore, the overall complexity of forward propagation and backpropagation in all edge devices is given by $\sum_{k=1}^{K}O(2m_k(dn+(N-1)n^2+Jn))=O(2m((N-1)n^2+(J+d)n))$ \cite{lecun2015deep}.

As summarized in Table \ref{4schems_comp}, for LoLaFL, if we only focus on the dominant part (i.e., terms with $d^3$) in the expressions, the HM-like and CM-based schemes have comparable computational complexity. 
The computational complexity of LoLaFL is proportional to $d^3$ and $J$, while for traditional FL, the dominant part is proportional to $n^2$ and $N$. This indicates that the bottleneck of LoLaFL is primarily related to the complexity of the datasets, while that of traditional FL is associated with the width and depth of the neural network. Additionally, it is observed that the computational complexity of LoLaFL scales linearly with respect to the number of devices $K$, and the scaling coefficient primarily arises from the matrix inversion and SVD in the two schemes, respectively. We acknowledge that this could potentially become a bottleneck as the number of edge devices significantly increases. In large-scale FL deployments (e.g., \(K>\!\!>100\)), the device selection strategy can be applied to alleviate the increased computational burden (see e.g., \cite{mcmahan2017communication}).

As demonstrated by the experiments in the sequel, the CM-based scheme achieves over $97\%$ reduction in total latency (communication latency and computation latency) compared with traditional FL. The low latency results from the following three aspects:
\begin{itemize}
    \item \textbf{Forward-only propagation}: In LoLaFL, the layers are constructed in a forward manner, and the parameters are calculated directly and deterministically according to formulae. Since these parameters of each layer in LoLaFL are near-optimal, once a layer is constructed, no BP is needed. In contrast, traditional FL requires random initialization and multiple rounds of BP to update the whole model.
    Therefore, we are comparing a layer in LoLaFL with the entire black-box model in traditional FL, in each communication round. 
    \item \textbf{Novel aggregation scheme}: Unlike HM-like aggregation and FedAvg, the novel CM-based aggregation makes use of CMs. The low-rank structures of features allow for compression of the CMs, enabling the transmission of a smaller volume of data (singular vectors and singular values rather than CMs). This helps to further reduce the communication latency.
    \item \textbf{Minimal communication round}: In our experiments, it has been observed that merely a few rounds of communication can achieve comparable accuracy. The reasons are twofold: model size and normalization. 1) Generally, in deep learning, a larger model size means better performance. ReduNet and ResNet have some similarities \cite{chan2022redunet}, and the parameter number of a single layer of ReduNet (about $6.8\times 10^6$, near-optimal) is already comparable with the entire ResNet-18 (about $1.1\times 10^7$, not optimal in the first communication round). 2) Regardless of the scale of the learning rate, the transformed features are always normalized, which facilitates training with a relatively large learning late. In contrast, in traditional FL, only an appropriate learning rate leads to good performance.
\end{itemize}

\subsection{Privacy Guarantee}\label{privacy}
In traditional FL, the original data are kept locally and are not sent to the server, thereby ensuring data privacy. In LoLaFL, although the original data remain local, the transmitted parameters are related to features that are transformed from the original data. We will demonstrate that, for both the HM-like and CM-based schemes in LoLaFL, it is not possible to derive the features from the transmitted parameters, let alone recovering the original data. The details are as follows.

Let $\mathbf{Z}_{\ell,k}^j$ be the features belonging to class $j$ in layer $\ell$ at edge device $k$. For the edge server, even if it can get the CMs by either using \eqref{Cl_jk} as
\begin{equation}
    \mathbf{Z}_{\ell,k}^j \mathbf{Z}_{\ell,k}^{j*}= \mathbf{Z}_{\ell,k}\mathbf{\Pi}_{k}^j\mathbf{Z}_{\ell,k}^*=(\mathbf{C}_{\ell}^j)^{-1}-({1}/{\alpha_k^j})\mathbf{I}\nonumber
\end{equation}
(for the HM-like scheme) or receiving directly (for the CM-based scheme), it cannot recover the original features $\mathbf{Z}_{\ell,k}^j$ from the CMs, and the reasoning is as follows.
Denote $\mathbf{Y}\triangleq\mathbf{Z}_{\ell,k}^j\mathbf{Z}_{\ell,k}^{j*}$, as the calculated/received positive semi-definite matrix. Indeed we can find a solution $\mathbf{Z}_0$ for equation $\mathbf{Z}\mathbf{Z}^*=\mathbf{Y}$ which satisfies $\mathbf{Z}_0\mathbf{Z}_0^*=\mathbf{Y}$ (e.g., by Cholesky factorization \cite{moon2000mathematical}). But for any orthogonal matrix $\mathbf{Q}$, $\mathbf{Z}_{1}=\mathbf{Z}_0\mathbf{Q}$ is also a solution for equation $\mathbf{Z}\mathbf{Z}^*=\mathbf{Y}$ because $\mathbf{Z}_{1}\mathbf{Z}_{1}^*=\mathbf{Z}_0\mathbf{Q}\mathbf{Q}^*\mathbf{Z}_0^*=\mathbf{Y}$. Therefore, the solution is not unique unless other constraints are provided, which means the original features $\mathbf{Z}_{\ell,k}^j$ cannot be derived.

Still, there is one exception where original $\mathbf{Z}_{\ell,k}^j$ can be obtained exactly, i.e., the sample number of some classes in some devices is only $1$. In this situation, we can obtain the original data $\mathbf{Z}_{\ell,k}^j=[\sqrt{\mathbf{Y}(1,1)}, \ \sqrt{\mathbf{Y}(2,2)}, \ \cdots , \ \sqrt{\mathbf{Y}(d,d)}]^*$ from the received $\mathbf{Y}$. Here the aforementioned orthogonal matrix $\mathbf{Q}$ degenerates into a number, namely one, hereby resulting the unique solution. However, we can safely ignore this exception, if we assume that there are no devices where there is only one sample belonging to a certain class.

For an edge device, it can only obtain the CMs of the other edge device when $K=2$, by subtracting its own local CMs from the global CMs which are either calculated (for the HM-like scheme) or received (for the CM-based scheme). However, even in this scenario, the original features of the other device cannot be derived due to the same reasoning as for the edge server. Consequently, we conclude that both proposed schemes provide a privacy guarantee.

Although we have mathematically demonstrated that deriving features using covariance matrices is not possible under certain conditions, we acknowledge that when the number of samples is extremely limited, it may be feasible to reconstruct approximate features (see, e.g., \cite{zhu2019deep}). Furthermore, the risk of membership inference, which can determine whether a specific data point belongs to the training dataset, may arise \cite{shokri2017membership}. We propose that differential privacy (DP), which adds noise (e.g., Gaussian or Laplacian) to the transmitted parameters, could mitigate these risks. As for the Gaussian noise injection, a particularly promising approach involves considering analog transmission and implementing AirComp, which transforms the distortions introduced by the wireless channel into a privacy-preserving mechanism (see, e.g., \cite{park2023differential}). Additionally, converting hard labels into soft labels in the training process could further reduce the risk of membership inference. Specifically, taking the membership matrix $\mathbf{\Pi}^j$ and a sample $i$ as an example, we do not require $\mathbf{\Pi}^j(i,i)=1$ or $\mathbf{\Pi}^j(i,i)=0$. Instead, we allow $\mathbf{\Pi}^j(i,i)\in[0,1]$ as long as $\sum_{j=1}^J \mathbf{\Pi}^j(i,i)=1$.\footnote{Though the proposed methods can mitigate the privacy issue, the accuracy may degrade. How to realize the best tradeoff between privacy and accuracy falls out of the scope of this paper. We advocate that this could be a potential direction in the future work.}

\section{Experimental Results}\label{simulation}

\subsection{Experimental Settings}

\begin{itemize}
    \item \textbf{Communication setting:} We consider a FL system comprising an edge server and $K=10$ edge devices. 
    The available frequency spectrum with bandwidth $B=10 \ \textrm{MHz}$ is divided into $M=K$ orthogonal subchannels.
    The threshold for truncated channel inversion is set as $\tau=0.105$, which corresponds to an outage probability of about $\xi=0.1$. Device $k$ uploads local parameters only when $|h_k|^2\geq\tau$, otherwise it quits parameters uploading in this communication round. 
    To guarantee a high quantization resolution, we set $Q=32$ \cite{liu2023over}. 
    \item \textbf{Metrics:} The test accuracy and total latency are two important metrics used to compare the performance of LoLaFL and traditional FL. The former utilizes the test set to assess the model at each learning stage, indicating how well the model is trained and its ability to generalize. The specific definition of the latter metric is given by
    \begin{equation}
    T_{\mathrm{total}} = \sum_{\ell=1}^{L'}\max_{k\in \mathcal{K}}\{T_{\mathrm{comm},\ell,k}+T_{\mathrm{comp},\ell,k}\},
\end{equation}
where $L'$ represents a given number of communication rounds, and $T_{\mathrm{comp},\ell,k}$ is the computation latency for device $k$ in communication round $\ell$. A total of $50$ IID experiments are conducted with different channel realizations, to yield the average performance. Specifically, we evaluate the performance at each communication round, averaging the test accuracy and total latency up to that round.
    \item \textbf{LoLaFL schemes:} The hyperparameters in LoLaFL schemes are as follows: $L=1$, $\eta=0.1$, $\epsilon=1$, $\lambda=500$. For LoLaFL with CM-based aggregation, $\beta_0=0.98$. Additionally, to provide a benchmark, we consider LoLaFL with the classic aggregation approach as in FedAvg, i.e., using arithmetic mean for aggregation, and denote it as LoLaFL (FedAvg).
    \item \textbf{Traditional FL schemes:} We implement traditional FL using ResNet-18, whose parameter number is approximately $W=1.1\times 10^7$ and the learning rate is set as $\eta=0.1$ \cite{he2016deep}. We consider two schemes, including the classic FedAvg\cite{mcmahan2017communication}, and the FedProx, which was proposed to deal with non-IID data distributions by adding a proximal term to the local loss to penalize the client for straying too far from the global model\cite{li2020federated}. We set the client selection probability to one for both schemes. For FedProx, we set the proximal term coefficient to $\mu=1$\cite{li2020federated}.
    \item \textbf{Real-world datasets:} Three popular datasets, MNIST, Fashion-MNIST, and CIFAR-100 are utilized in the experiments. Both the first two datasets consist of a training set containing $60,000$ labeled data samples and a test set of $10,000$ labeled data samples, each comprising 10 classes; while the last one consists of a training set containing $50,000$ labeled data samples and a test set of $10,000$ labeled data samples, comprising 100 classes. The MNIST dataset consists of handwritten digits ranging from 0 to 9, while the Fashion-MNIST and CIFAR-100 dataset include various objects such as trousers and airplane. The images in (Fashion-) MNIST datasets are grayscale and have a size of $28\times 28$, and consequently, we have $d=28\times 28=784$ and $J=10$ for both datasets. The images in CIFAR-100 dataset are colored and have a size of $32\times 32$. We have $d=3\times 32\times 32=3072$ and we use a subset of it with $J=3$. Each device is assigned $m_k=1,200$ labeled data samples from (Fashion-) MNIST dataset and $m_k=30$ labeled data samples from CIFAR-100 dataset for training. We consider both IID and non-IID settings for data partition and allocation. In the IID setting, each device randomly obtains $m_k$ labeled data samples from the training set. We consider two non-IID settings to comprehensively investigate the impacts of data heterogeneity, outlined as follows. Non-IID (a): $m_k\times K$ samples are initially selected at random from the training set, which are then sorted according to their respective classes and sequentially allocated to each device. This ensures that no device contains more than two classes \cite{mcmahan2017communication}. Non-IID (b): each device is randomly assigned a specific class and subsequently obtains $m_k$ labeled data samples belonging to that class from the training set. In this setting, each device contains only a single class, representing a more stringent condition \cite{zhao2018federated}. For testing, all available samples from the test set are used.

\end{itemize}

\subsection{Learning Performance of LoLaFL}

\begin{figure*}[h]
\centering
    \subfigure[MNIST]{\includegraphics[width=0.3\linewidth]{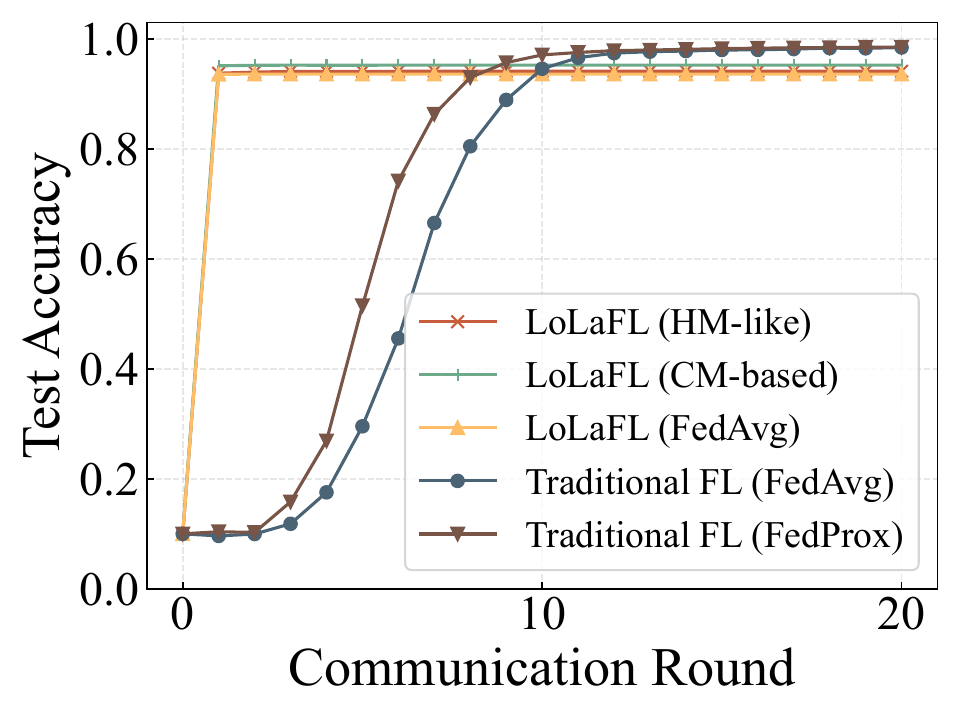}\label{acc_compare_whiteblack_epoch_dig.a}}
    \subfigure[Fashion-MNIST]{\includegraphics[width=0.3\linewidth]{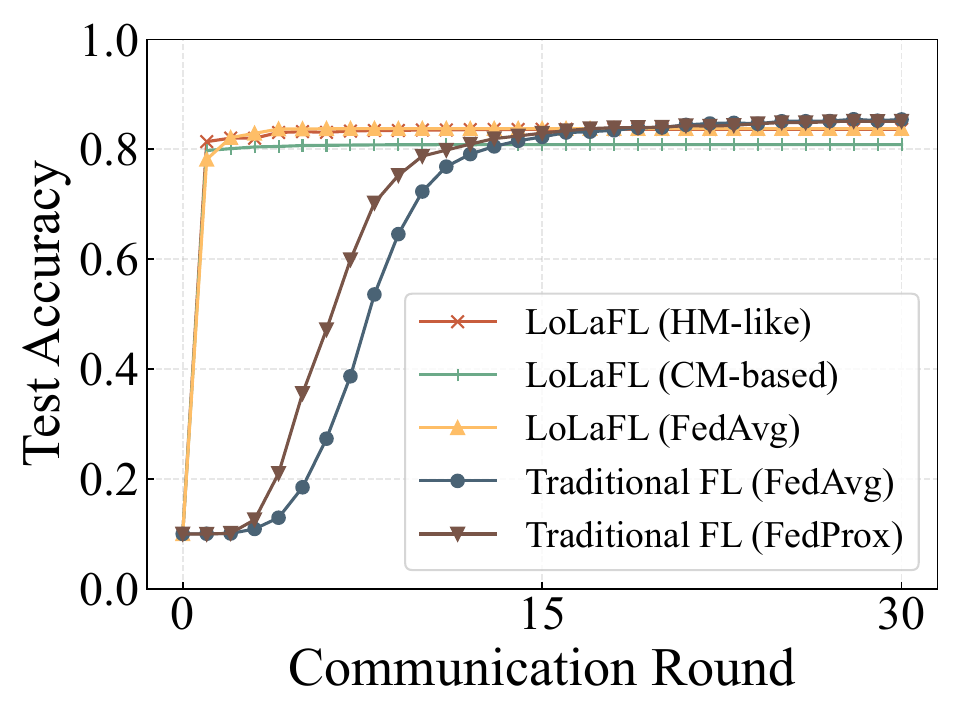}\label{acc_compare_whiteblack_epoch_dig.b}}  
    \subfigure[CIFAR-100 (3-class subset)]{\includegraphics[width=0.3\linewidth]{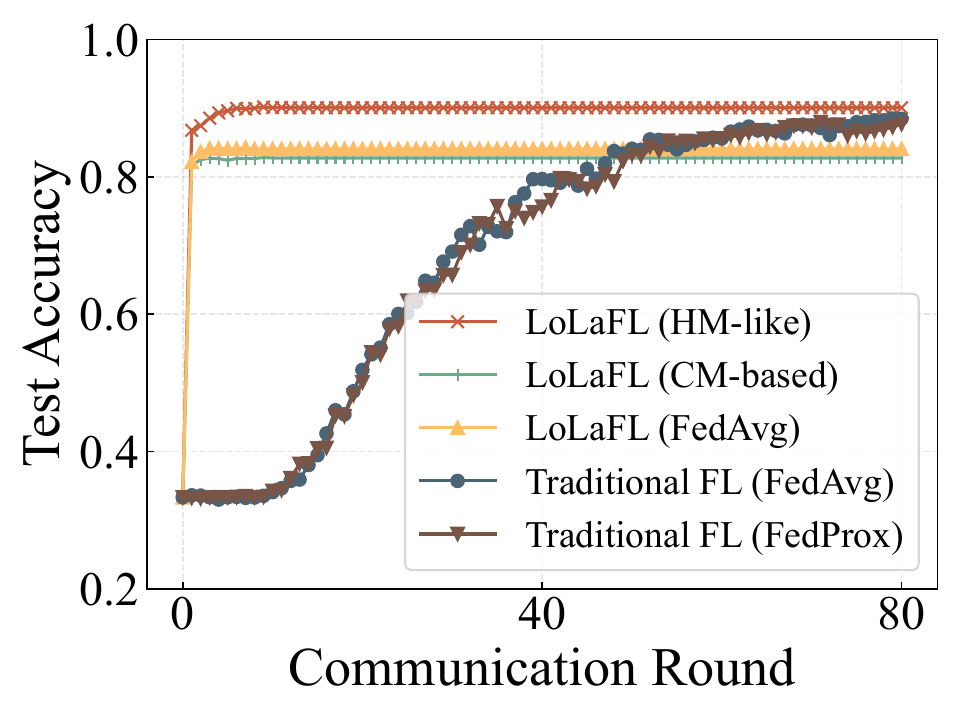}\label{acc_compare_whiteblack_epoch_dig.c}} 
    \caption{Learning performance comparison between LoLaFL and traditional FL w.r.t. communication round.}
    \captionsetup{justification=justified}
    \label{acc_compare_whiteblack_epoch_dig}
  \end{figure*}

\begin{figure*}[h]
\centering
        \subfigure[MNIST]{\includegraphics[width=0.3\linewidth]{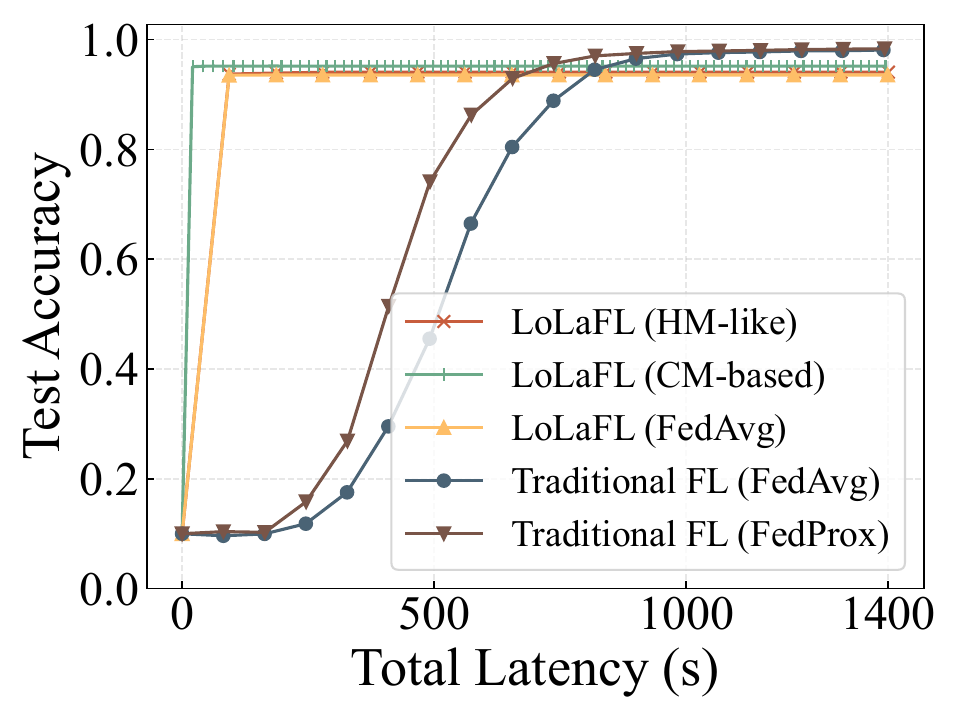}\label{acc_compare_whiteblack_total_dig.a}}
    \subfigure[Fashion-MNIST]{\includegraphics[width=0.3\linewidth]{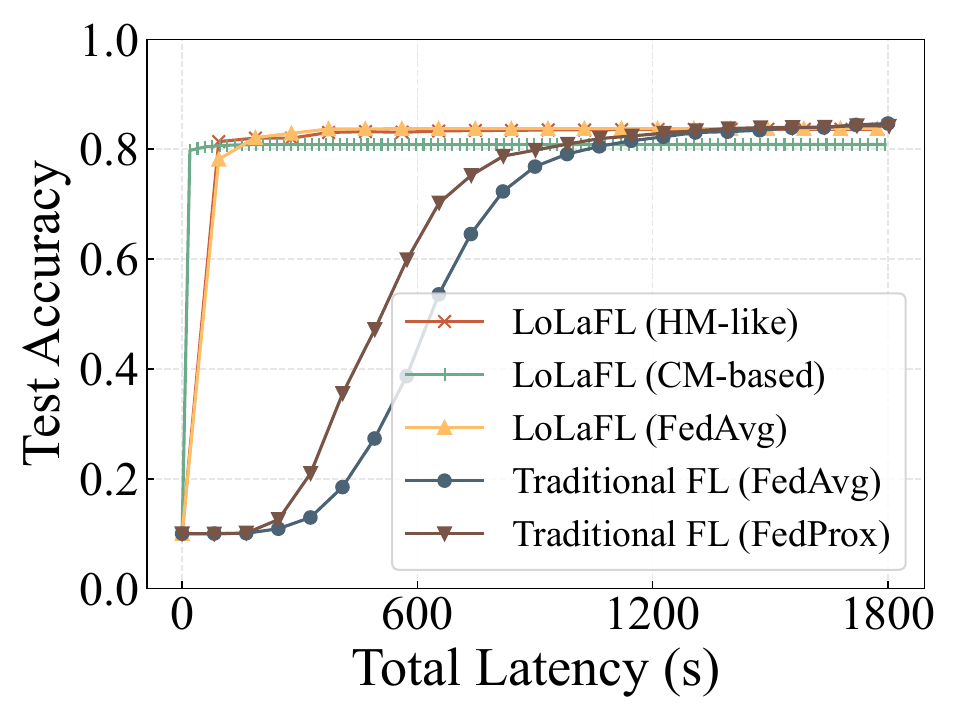}\label{acc_compare_whiteblack_total_dig.b}}  
    \subfigure[CIFAR-100 (3-class subset)]{\includegraphics[width=0.3\linewidth]{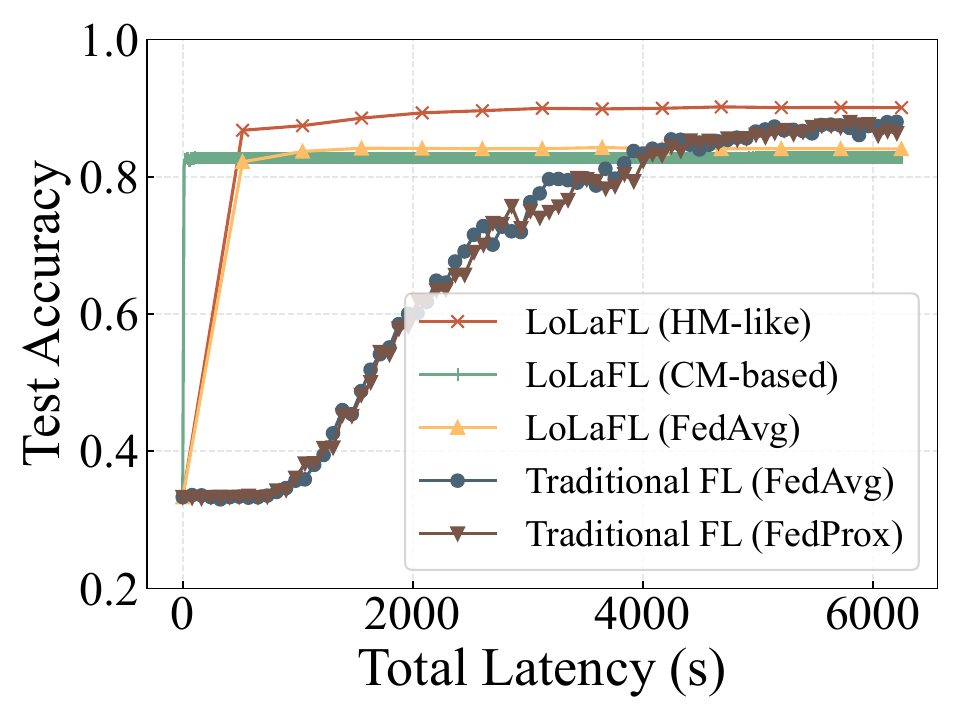}\label{acc_compare_whiteblack_total_dig.c}}  
    \caption{Learning performance comparison between LoLaFL and traditional FL w.r.t. total latency.}
    \captionsetup{justification=justified}
    \label{acc_compare_whiteblack_total_dig}
  \end{figure*}

The performance of LoLaFL and traditional FL is compared in Fig.~\ref{acc_compare_whiteblack_epoch_dig}-\ref{acc_compare_whiteblack_total_dig}.
We begin by examining the convergence characteristics of LoLaFL with different aggregation schemes. For the MNIST dataset, the three schemes for LoLaFL exhibit nearly identical increases in test accuracy as the number of layers increases, with LoLaFL with CM-based aggregation showing a slight advantage. However, for the more complex Fashion-MNIST and CIFAR-100 datasets, LoLaFL with HM-like aggregation generally achieves a higher accuracy, serving as the upper bound. Specifically, for CIFAR100, the rest two schemes have identical test accuracy over the communication rounds, but have a noticeable gap to the upper bound, due to the information loss resulting from the low-rank approximation employed in the CM-based scheme, and the non-optimal aggregation in the FedAvg scheme, respectively. For all datasets, LoLaFL with FedAvg nearly serves as the lower bound for LoLaFL with HM-like aggregation. Another important observation is that the accuracy of the white-box schemes has achieved a high level even in the first layer, while the subsequent layers contribute to a limited increase in accuracy. This observation motivates us to transmit only the first layer in LoLaFL and justifies why we set $L=1$ in the following experiments.

While traditional FL has the potential to outperform LoLaFL when given sufficient communication rounds, the required number of rounds and total latency are considerable. Taking traditional FL (FedProx) as an example, it outperforms traditional FL (FedAvg) for the MNIST and Fashion-MNIST datasets. However, it still needs around $10$ communication rounds to surpass LoLaFL for the MNIST and Fashion-MNIST datasets, and around $50$ rounds for the CIFAR-100 dataset.
This demonstrates the advantage of LoLaFL over traditional FL: to achieve comparable accuracy, LoLaFL requires only $1/10$ (or even less) of the communication rounds compared to traditional FL, suppressing the communication overhead between the edge device and the edge server. When considering the total latency required for comparable accuracy, the HM-like and CM-based schemes in LoLaFL require less than $13\%$ and $3\%$ of the total latency associated with traditional FL, respectively. Specifically, the CIFAR-100 dataset demonstrates the most significant performance gain of LoLaFL with CM-based aggregation: it only needs about $0.3\%$ of the total latency associated with traditional FL.

\subsection{Effects of Network Parameters}

\begin{figure*}[h]
\centering
        \subfigure[MNIST]{\includegraphics[width=0.3\linewidth]{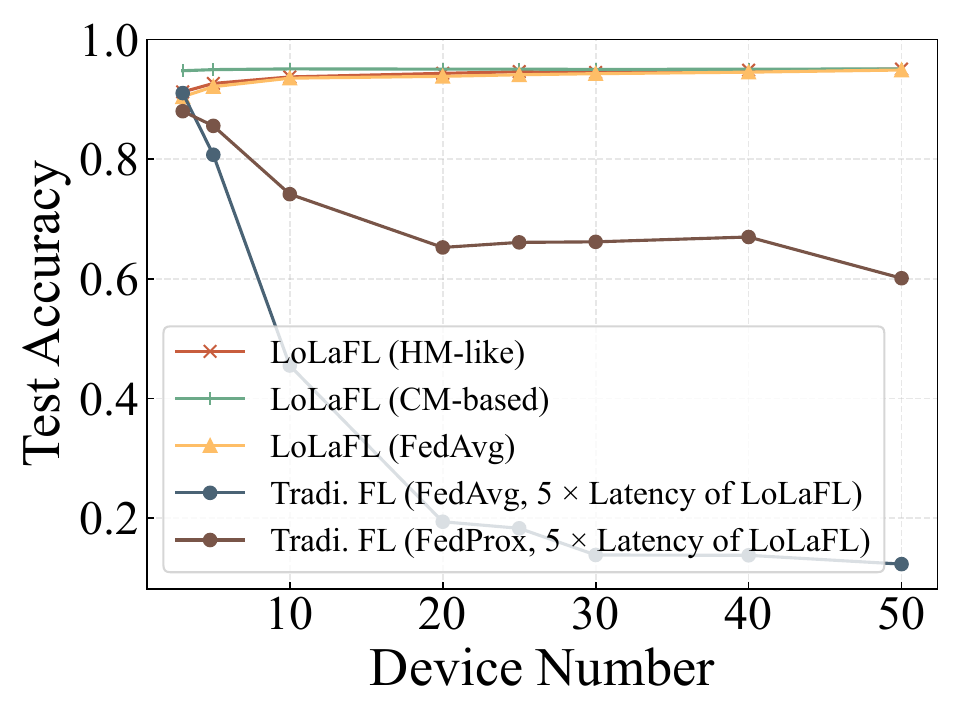}\label{acc_compare_whiteblack_device_num_dig.a}}
    \subfigure[Fashion-MNIST]{\includegraphics[width=0.3\linewidth]{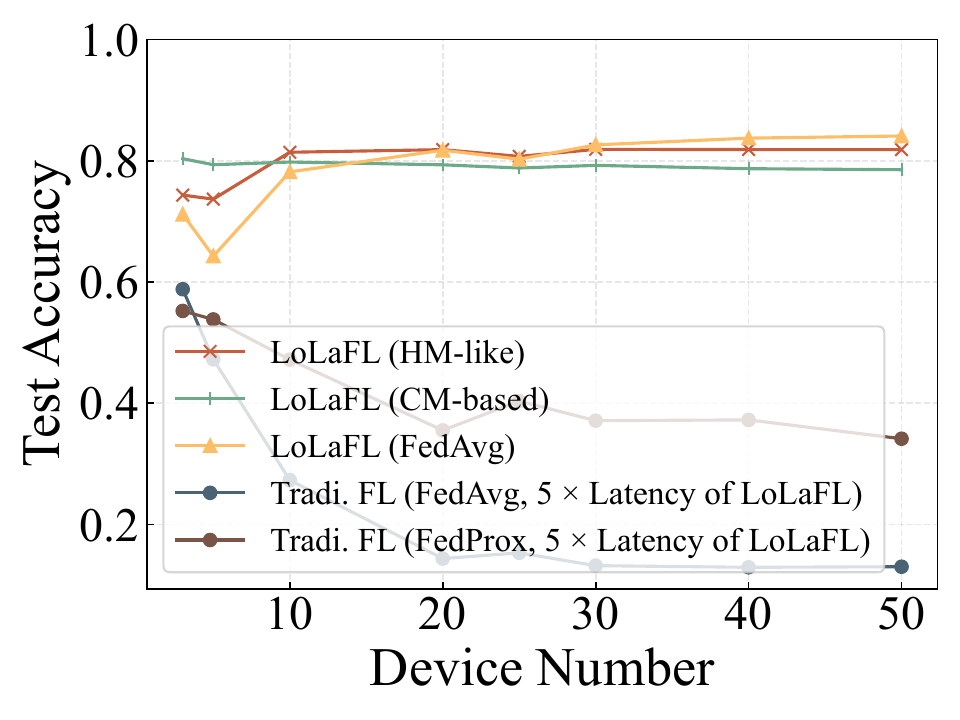}\label{acc_compare_whiteblack_device_num_dig.b}}  
    \subfigure[CIFAR-100 (3-class subset)]{\includegraphics[width=0.3\linewidth]{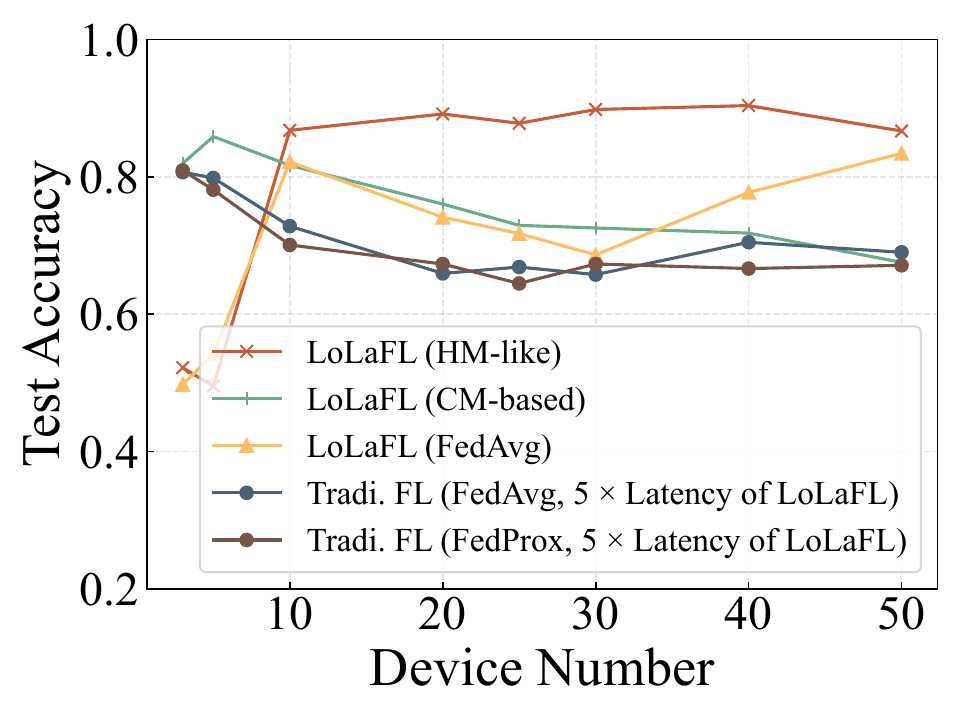}\label{acc_compare_whiteblack_device_num_dig.c}}  
    \caption{Learning performance comparison between LoLaFL and traditional FL w.r.t. device number.}
\label{acc_compare_whiteblack_device_num_dig}
    \captionsetup{justification=justified}
  \end{figure*}

\begin{figure*}[h]
\centering
    \subfigure[MNIST]{\includegraphics[width=0.3\linewidth]{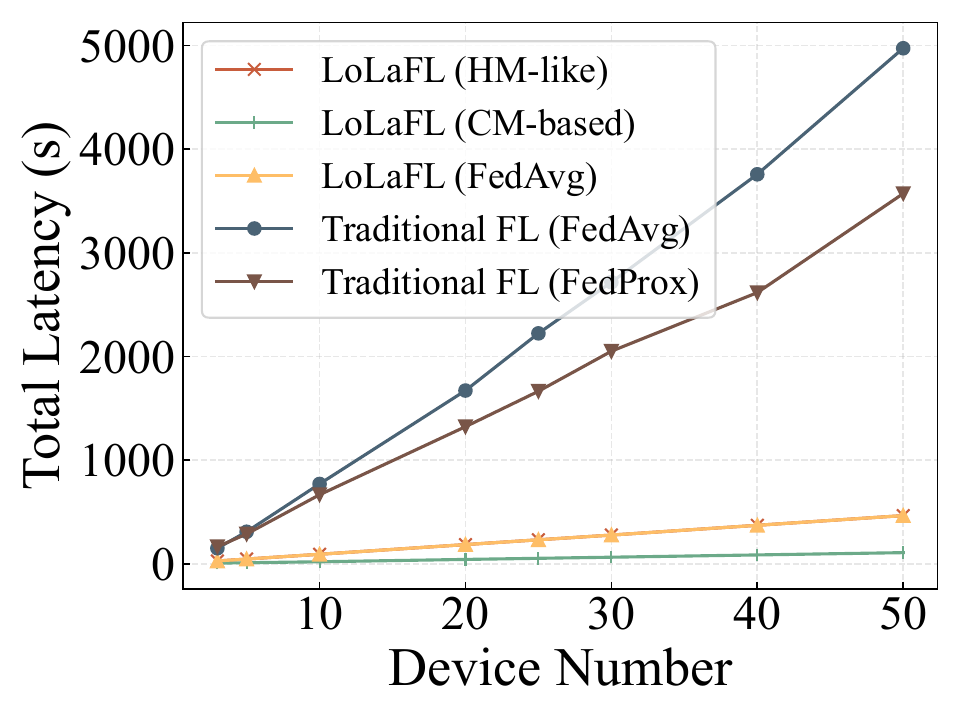}\label{latency_compare_whiteblack_device_num_dig.a}}
    \subfigure[Fashion-MNIST]{\includegraphics[width=0.3\linewidth]{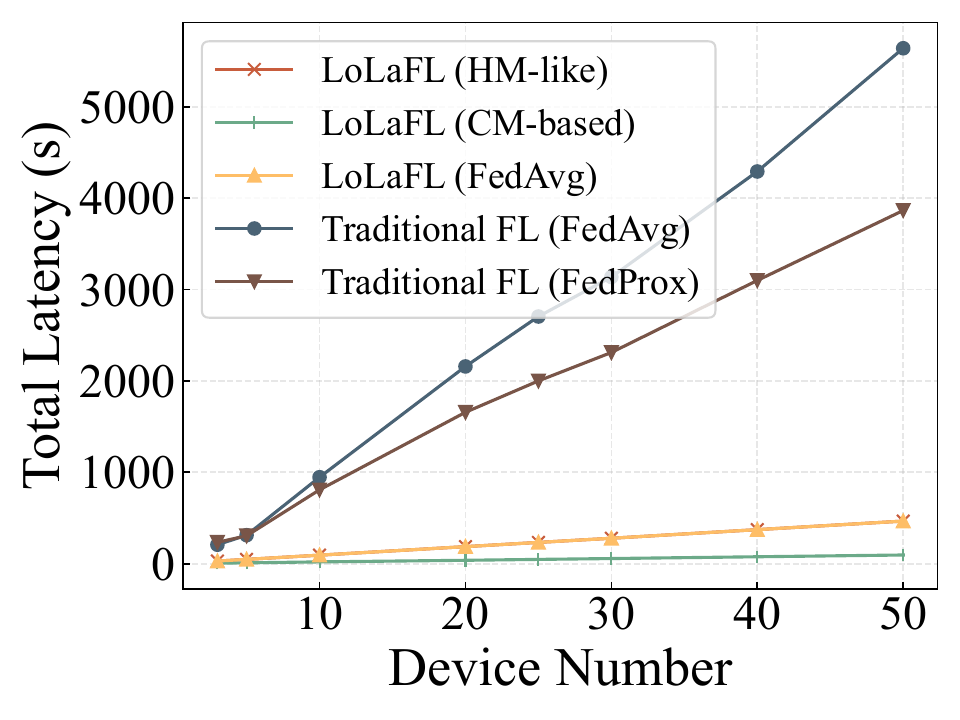}\label{latency_compare_whiteblack_device_num_dig.b}}  
    \subfigure[CIFAR-100 (3-class subset)]{\includegraphics[width=0.3\linewidth]{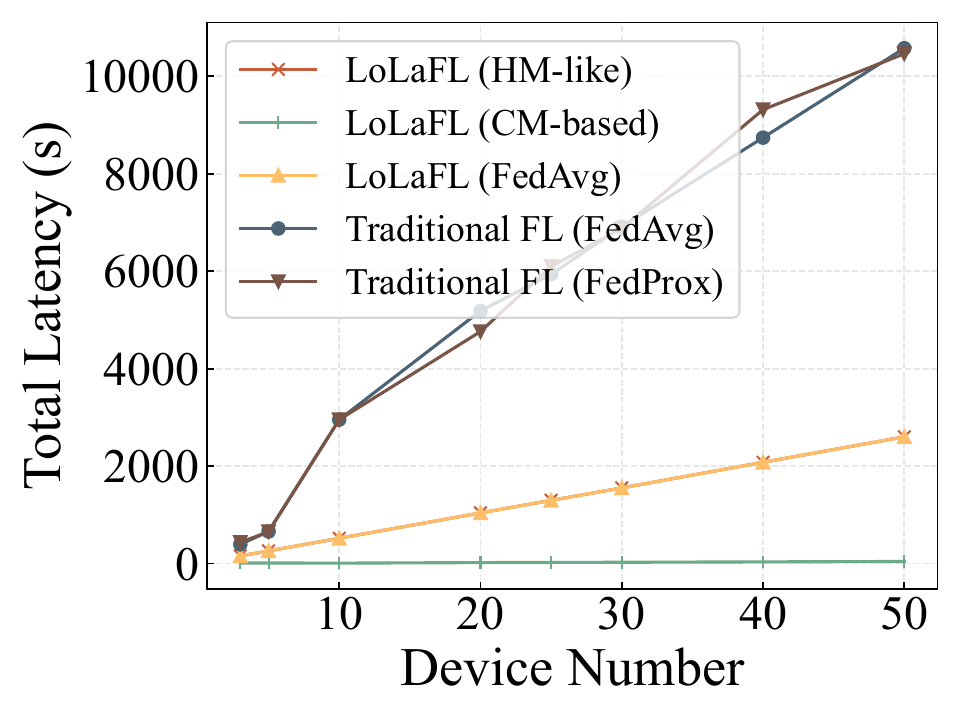}\label{latency_compare_whiteblack_device_num_dig.c}}  
    \caption{Total latency comparison between LoLaFL and traditional FL w.r.t. device number constrained by specific test accuracy. (a) Approximately $93\%$ test accuracy; (b) Approximately $76\%$ test accuracy; (c) Approximately $67\%$ test accuracy.}
    \captionsetup{justification=justified}
    \label{latency_compare_whiteblack_device_num_dig}
  \end{figure*}

We investigate the effects of two important network parameters, i.e., the device number and the outage probability on LoLaFL. 
With the current data allocation setting, more devices mean more samples are available for training. Fig.~\ref{acc_compare_whiteblack_device_num_dig} indicates the test accuracy of LoLaFL is better than traditional FL almost in all numbers of devices and datasets, with less than $1/5$ of the total latency. This is because LoLaFL directly calculates the NN parameters from the features, which captures the inherent structures of the data and features. There is a tendency for the performance of LoLaFL with HM-like aggregation and FedAvg to improve as the number of devices increases. However, the opposite trend is observed for the CM-based scheme. This may be attributed to the accumulated distortion caused by the low-rank approximation in the CM-based scheme. For traditional FL, especially for MNIST and Fashion-MNIST datasets, as the number of devices increases, the convergence speed is heavily affected, resulting in poor performance even when the total latency is $5$ times greater than that of LoLaFL. Although the available training samples increase, in traditional FL, local training during each communication round causes deviations of local models from the global model, and this phenomenon is exacerbated by the increasing number of devices, resulting in the performance degradation of traditional FL.
Fig.~\ref{latency_compare_whiteblack_device_num_dig} illustrates the total latency required to achieve satisfactory test accuracy across different schemes, where it is ensured that the test accuracy of traditional FL does not exceed that of LoLaFL at any number of devices. As the number of participating devices increases, a high-latency network emerges, with the increased total latency primarily resulting from the reduced bandwidth allocated to each edge device. Although the total latency for all three schemes generally increases linearly with device number, the rate of change for traditional FL is significantly steeper than that of LoLaFL. Consequently, traditional FL requires greater latency to achieve performance comparable with that of LoLaFL. 
The results indicate that traditional FL is not suitable for such scenarios, compared with LoLaFL.

\begin{figure*}[h]
\centering
    \subfigure[MNIST]{\includegraphics[width=0.3\linewidth]{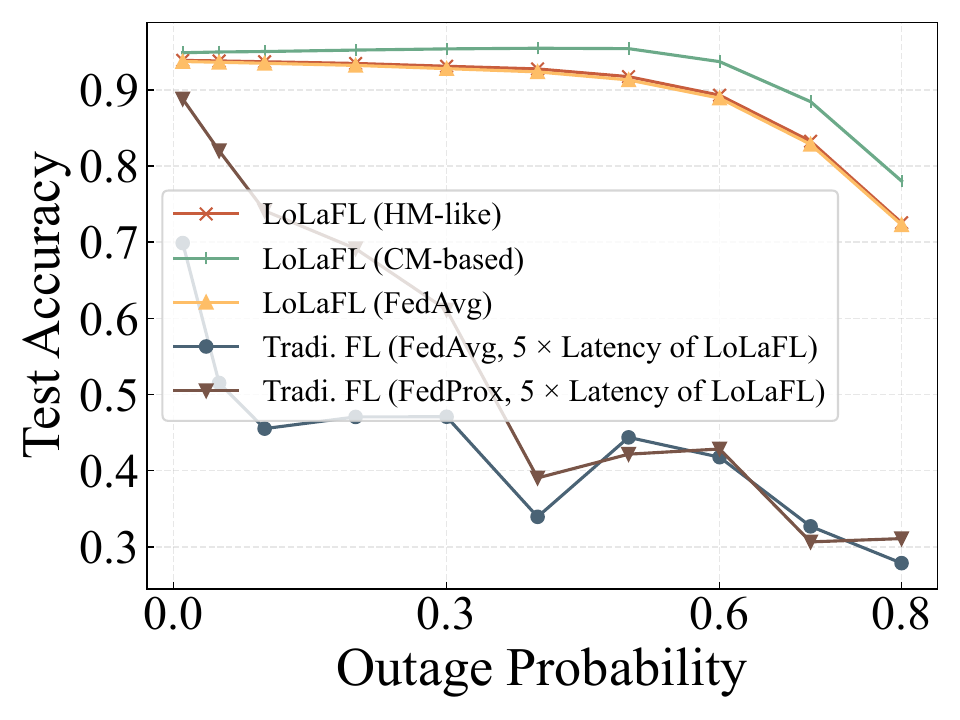}\label{acc_compare_whiteblack_outage_dig.a}}
    \subfigure[Fashion-MNIST]{\includegraphics[width=0.3\linewidth]{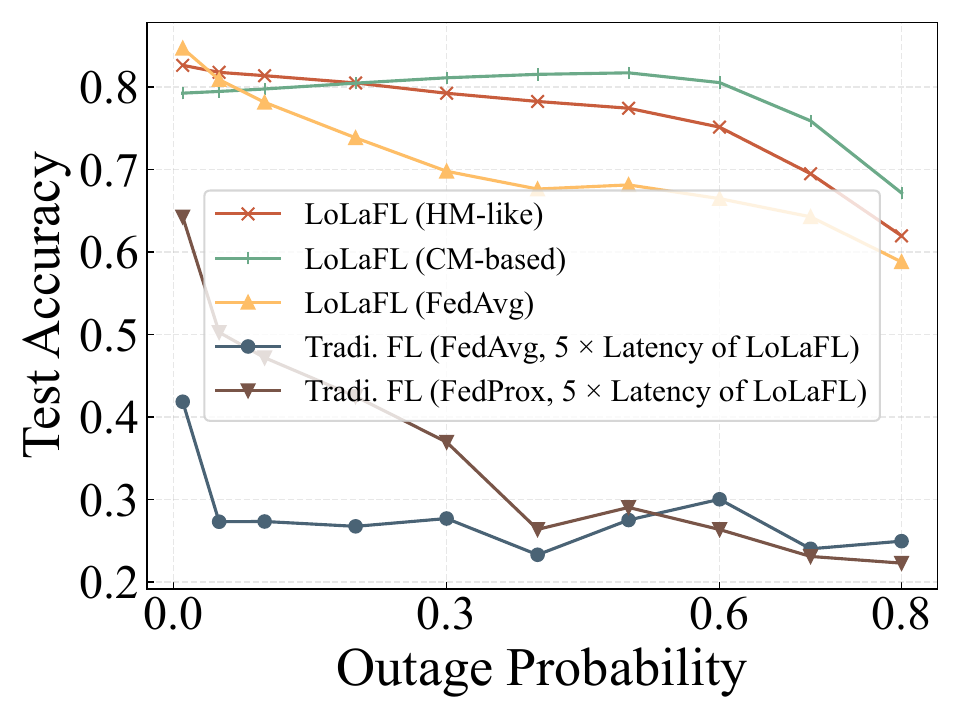}\label{acc_compare_whiteblack_outage_dig.b}}  
    \subfigure[CIFAR-100 (3-class subset)]{\includegraphics[width=0.3\linewidth]{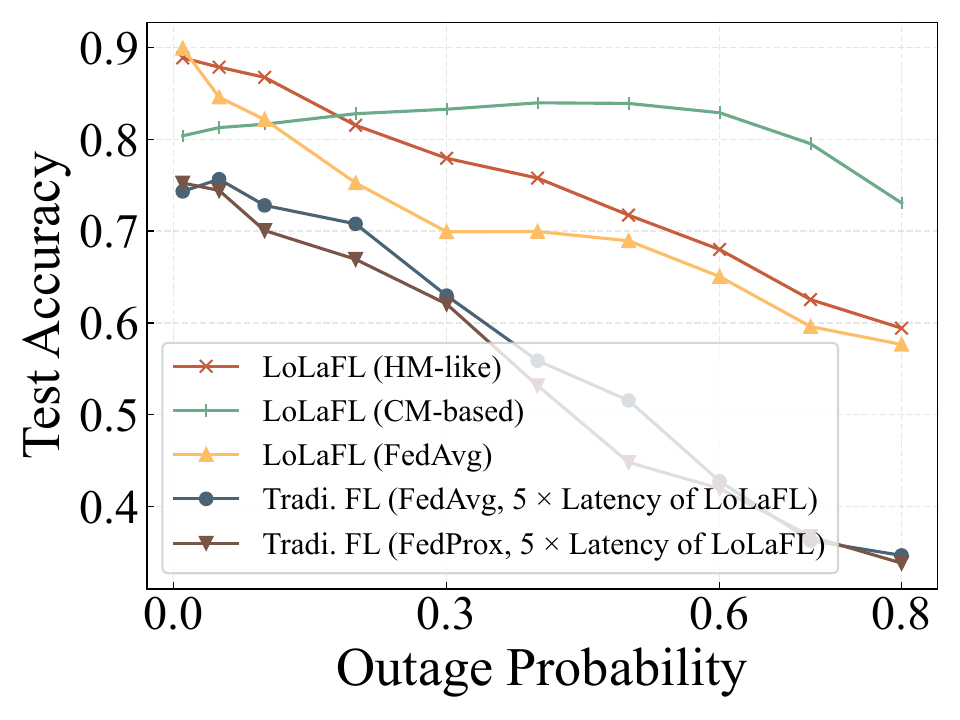}\label{acc_compare_whiteblack_outage_dig.c}}  
    \caption{Learning performance comparison between LoLaFL and traditional FL w.r.t. outage probability.}
    \captionsetup{justification=justified}
    \label{acc_compare_whiteblack_outage_dig}
  \end{figure*}

Fig.~\ref{acc_compare_whiteblack_outage_dig} illustrates the impacts of outage probability on different schemes, demonstrating how different schemes perform under different channel conditions. Additionally, since the total number of devices is fixed, varying outage probability reflects varying levels of device participation. The curves indicate that, the performance of LoLaFL with CM-based aggregation degrades when the outage probability exceeds approximately $0.5$. This observation can be attributed to the characteristics of white-box NN: although outages result in a reduction of available samples for training, only a portion of the training samples is sufficient for accurately constructing the NN parameters for LoLaFL. For the remaining two schemes for LoLaFL, although their performance degradation is more significant than that of LoLaFL with CM-based aggregation, they still outperform traditional FL across all outage probabilities, achieving this while utilizing only $1/5$ of the total latency of traditional FL.
Traditional FL is affected in all datasets, even when the outage probability is below $0.1$. This is because, in traditional FL, device outages can result in biased gradient estimations, leading to inefficient model training, which slows convergence and degrades performance.

\subsection{Compression of Covariance Matrices}

\begin{figure*}[h]
\centering
    \subfigure[MNIST]{\includegraphics[width=0.3\linewidth]{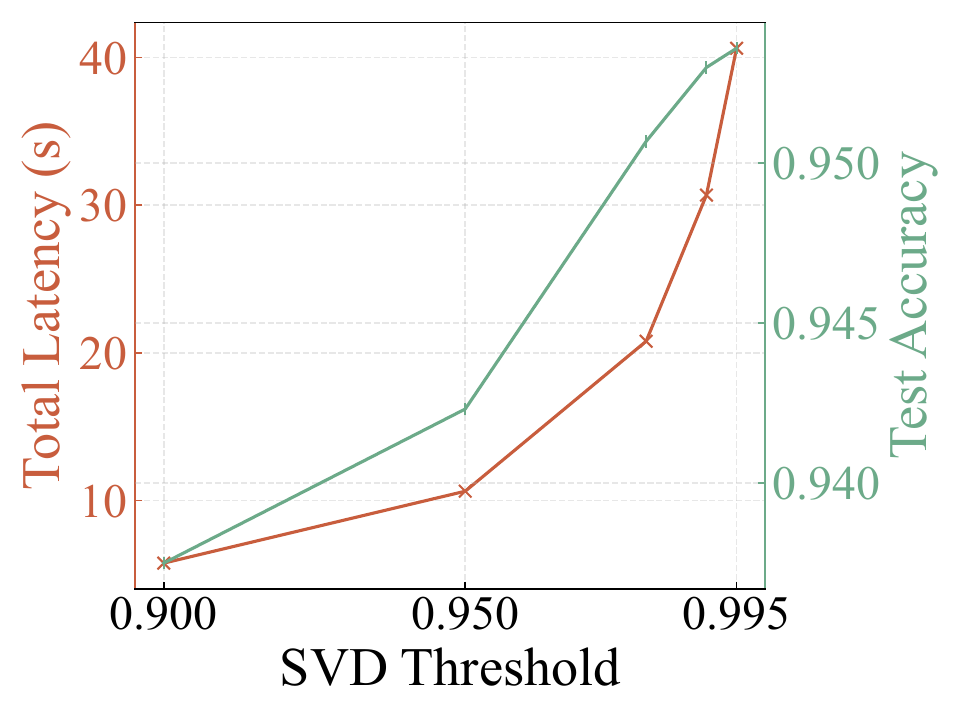}\label{acc_latency_compare_white_SVD_dig.a}}
    \subfigure[Fashion-MNIST]{\includegraphics[width=0.3\linewidth]{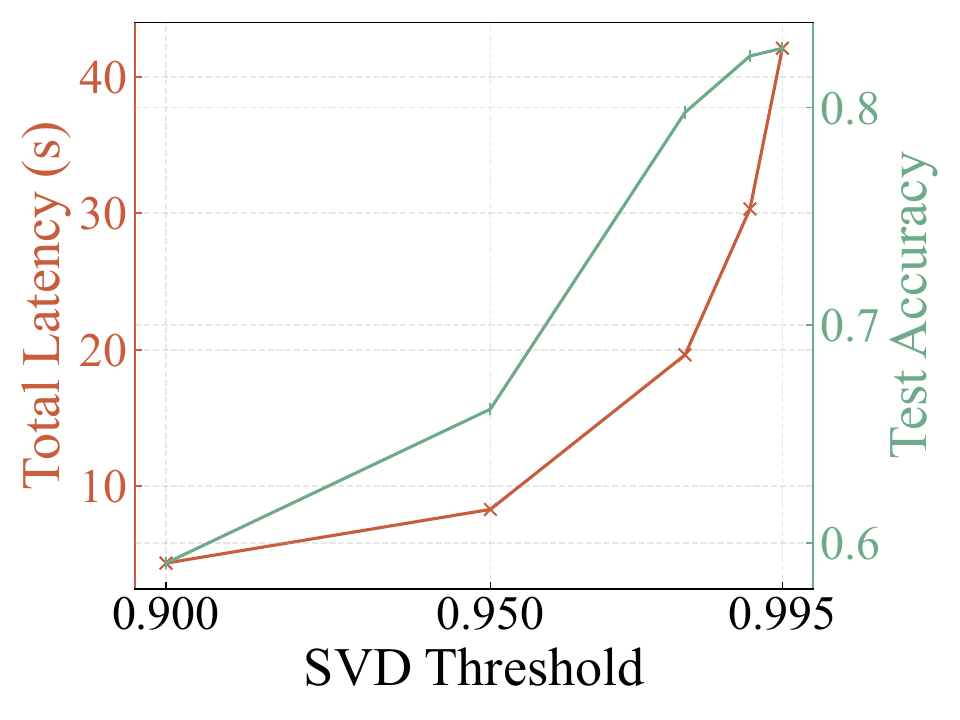}\label{acc_latency_compare_white_SVD_dig.b}}  
    \subfigure[CIFAR-100 (3-class subset)]{\includegraphics[width=0.3\linewidth]{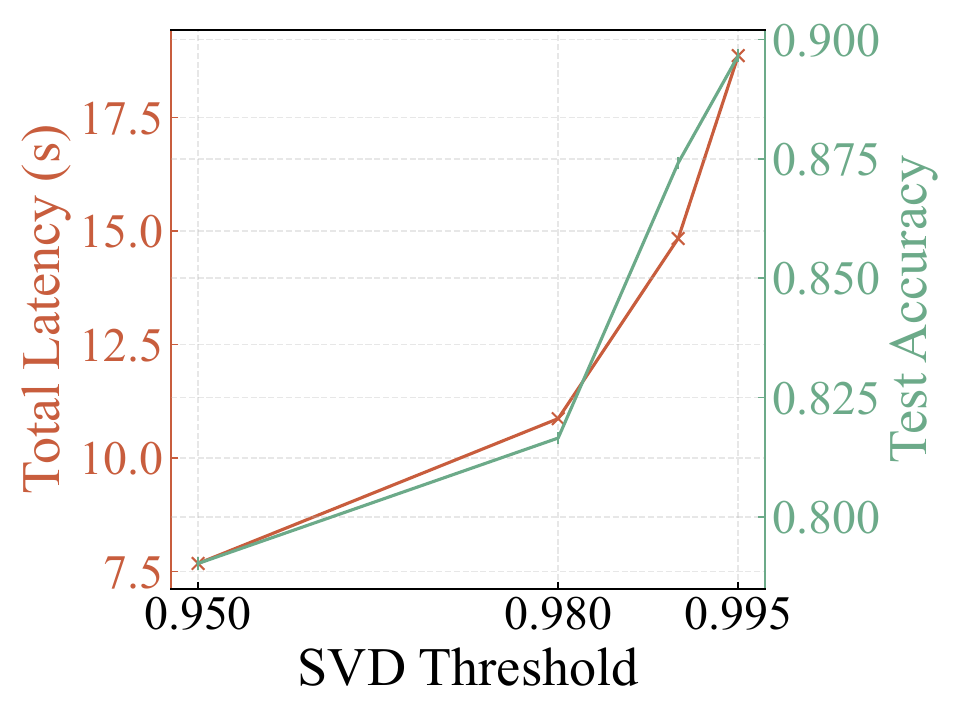}\label{acc_latency_compare_white_SVD_dig.c}}  
    \caption{The effects of SVD threshold on LoLaFL with CM-based aggregation.}
    \captionsetup{justification=justified}
    \label{acc_latency_compare_white_SVD_dig}
  \end{figure*}

We investigate how the SVD threshold influences the total latency and test accuracy for LoLaFL with CM-based aggregation, as shown in Fig.~\ref{acc_latency_compare_white_SVD_dig}. Theoretically, a higher SVD threshold permits the transmission of more singular vectors and values, which results in two key effects: 1) an increase in communication overhead, thereby increasing the total latency, and 2) a reduction in information loss within the reconstructed parameters, leading to improved learning performance. The curves presented in Fig.~\ref{acc_latency_compare_white_SVD_dig} agree with these expectations. This justifies our choice of setting the threshold to $0.98$ in our experiments, for the sake of achieving the best trade-off between accuracy and latency.

\subsection{IID and Non-IID}

\begin{figure}[h]
\centering
    \subfigure[MNIST]{\includegraphics[width=0.49\linewidth]{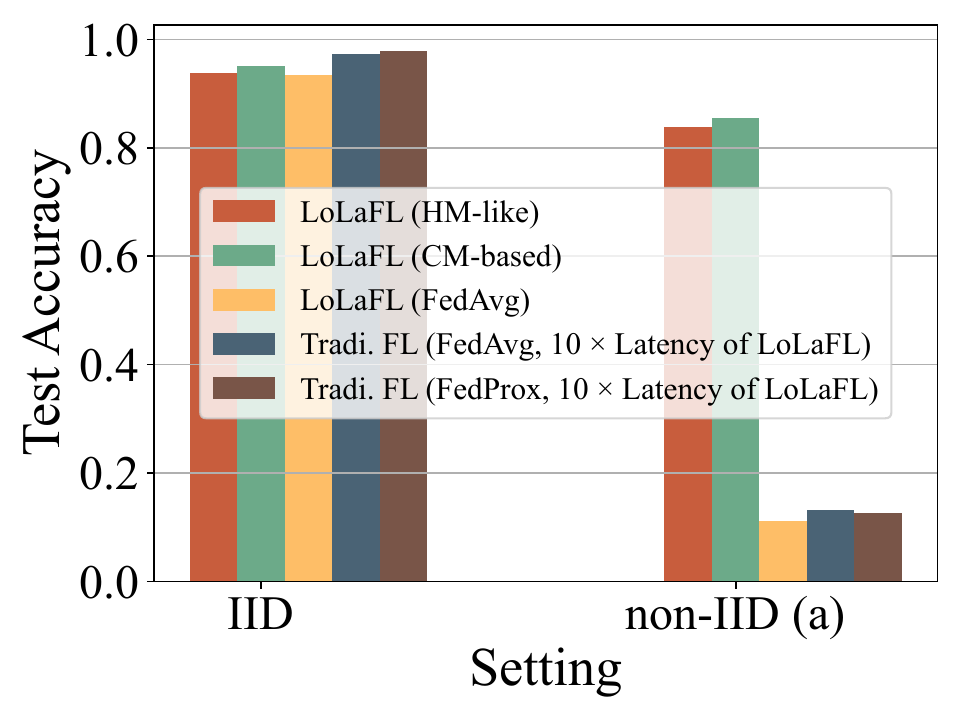}\label{acc_compare_whiteblack_latency_noniid_1.a}}
    \subfigure[Fashion-MNIST]{\includegraphics[width=0.49\linewidth]{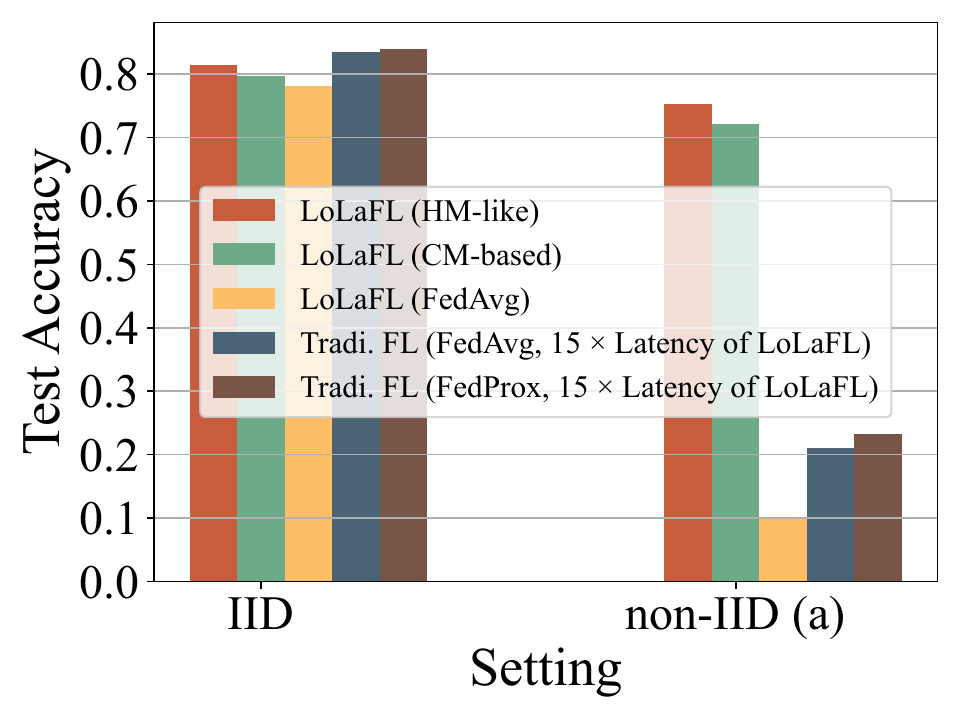}\label{acc_compare_whiteblack_latency_noniid_1.b}}  
    \subfigure[CIFAR-100 (3-class subset)]{\includegraphics[width=0.49\linewidth]{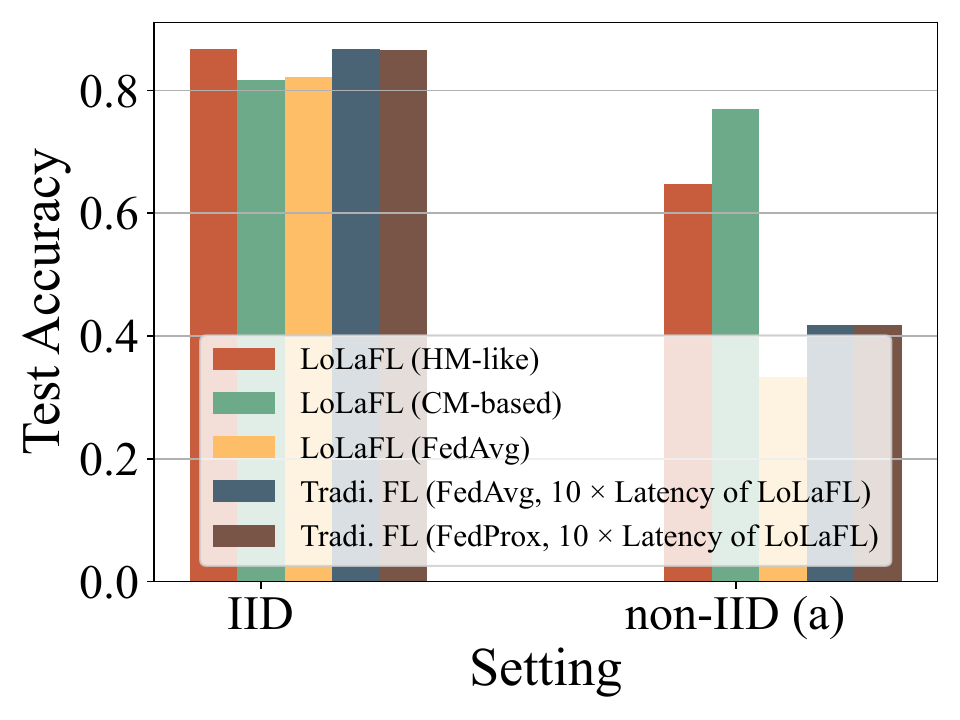}\label{acc_compare_whiteblack_latency_noniid_1.c}}  
    \caption{Learning performance comparison between LoLaFL and traditional FL on IID and non-IID (a) settings.}
    \captionsetup{justification=justified}
    \label{acc_compare_whiteblack_latency_noniid_1}
  \end{figure}

\begin{figure}[!t]
\centering
    \subfigure[MNIST]{\includegraphics[width=0.49\linewidth]{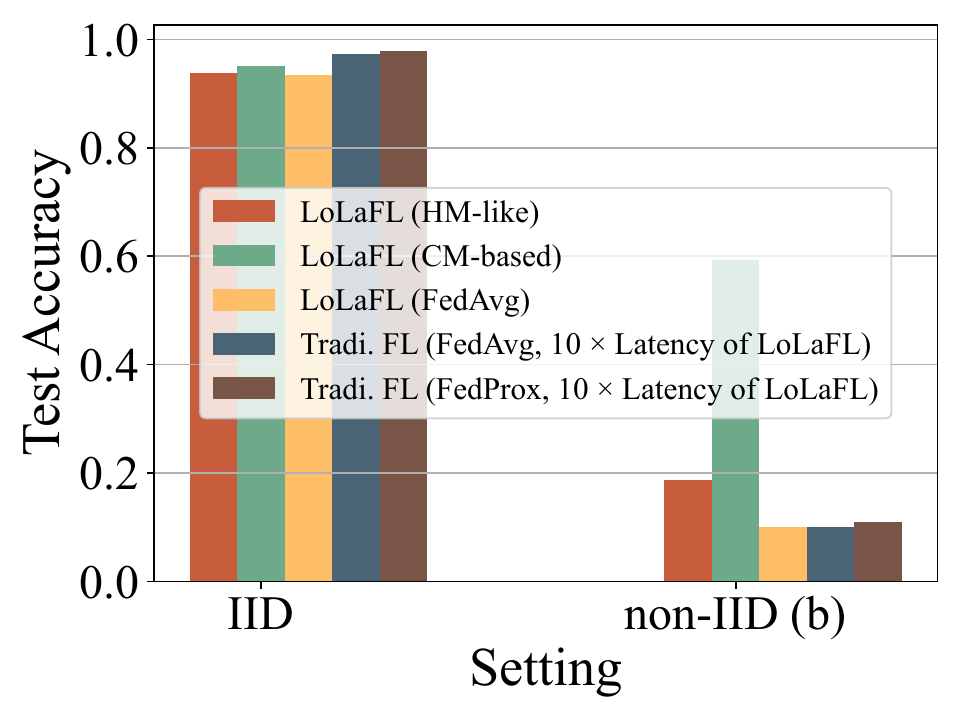}\label{acc_compare_whiteblack_latency_noniid_5.a}}
    \subfigure[Fashion-MNIST]{\includegraphics[width=0.49\linewidth]{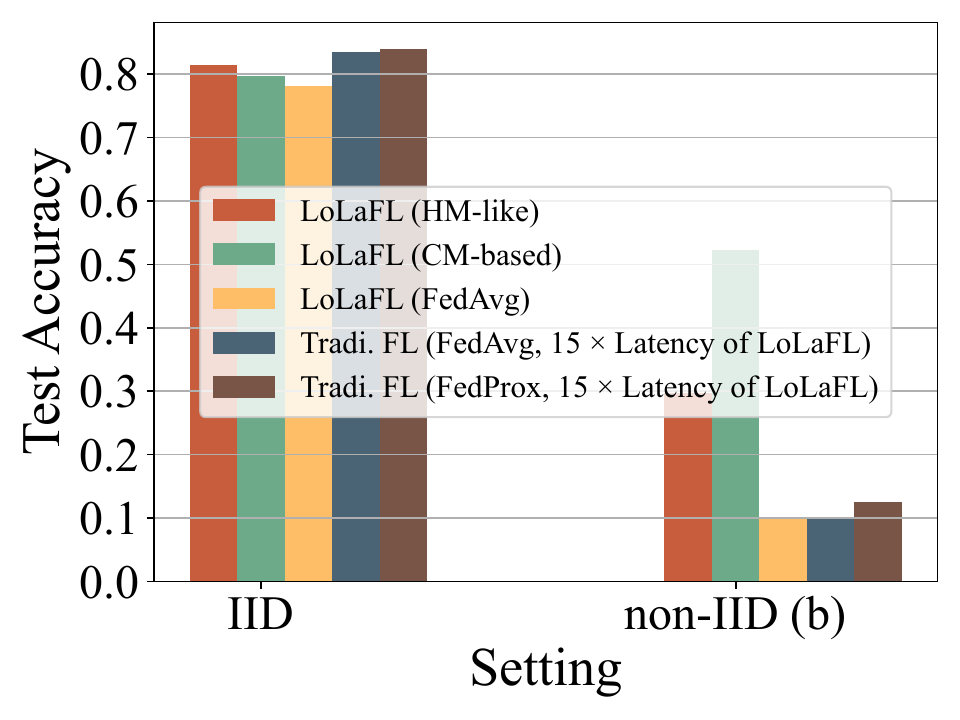}\label{acc_compare_whiteblack_latency_noniid_5.b}}  
    \subfigure[CIFAR-100 (3-class subset)]{\includegraphics[width=0.49\linewidth]{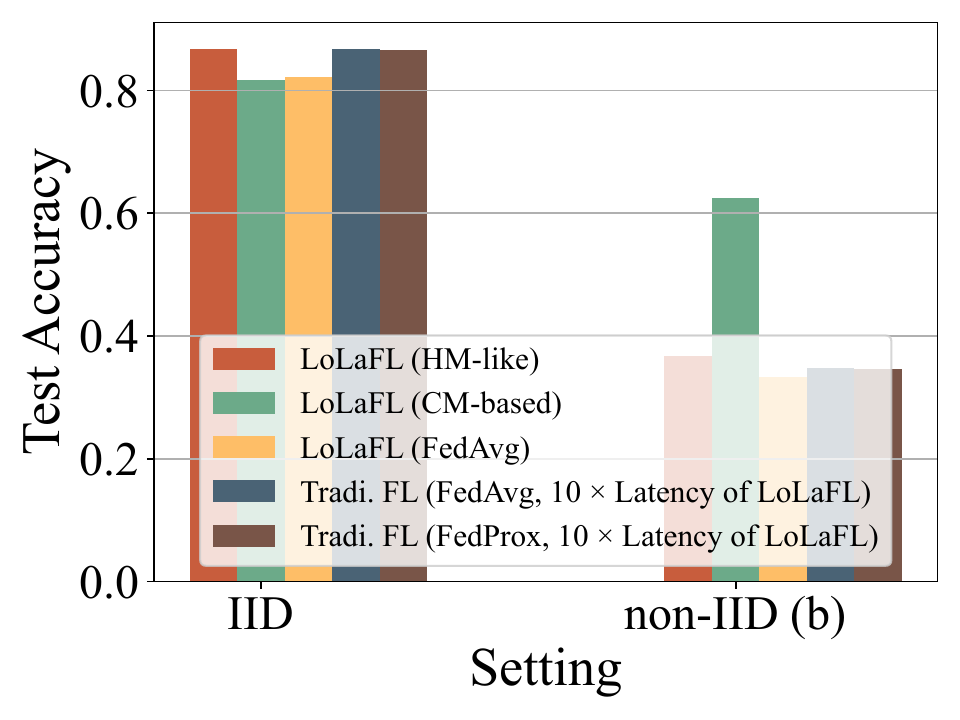}\label{acc_compare_whiteblack_latency_noniid_5.c}}  
    \caption{Learning performance comparison between LoLaFL and traditional FL on IID and non-IID (b) settings.}
    \captionsetup{justification=justified}
    \label{acc_compare_whiteblack_latency_noniid_5}
  \end{figure}

We investigate the influence of the non-IID data distributions on LoLaFL and traditional FL, as shown in Fig.~\ref{acc_compare_whiteblack_latency_noniid_1} and Fig.~\ref{acc_compare_whiteblack_latency_noniid_5}. In Fig.~\ref{acc_compare_whiteblack_latency_noniid_1}, the non-IID (a) has a minor influence on LoLaFL with HM-like and CM-based aggregations, while it significantly affects LoLaFL with FedAvg and traditional FL. 
When the data are IID, the local parameters across different devices may exhibit limited variation, allowing FedAvg to work for LoLaFL. However, since FedAvg is not the optimal aggregation scheme for LoLaFL, as demonstrated in Proposition 1, it cannot assist LoLaFL in addressing non-IID data. Compared with traditional FL, the NN parameters in LoLaFL are calculated from features directly, which means that the results remain consistent regardless of how the data are distributed across different devices, assuming the distortion induced by the channel is ignored. In contrast, for traditional FL, the heterogeneous data across different devices exacerbate the deviation of local models from the global model, leading to slower convergence and degraded performance. This demonstrates the substantial advantage of the proposed LoLaFL when dealing with non-IID data.
However, when considering the more stringent non-IID (b) setting in Fig.~\ref{acc_compare_whiteblack_latency_noniid_5}, the performance of all schemes is significantly impacted, even for LoLaFL (HM-like). In this extreme data heterogeneity scenario, LoLaFL (CM-based) emerges as the most effective scheme among the five schemes.

\section{Concluding Remarks}\label{concluding remarks}

In this paper, we have proposed the use of the state-of-the-art white-box approach to develop the novel LoLaFL framework accompanied by two novel nonlinear aggregation schemes. Compared with traditional FL, LoLaFL with HM-like and CM-based aggregations demonstrate tenfold and hundredfold reductions in latency, respectively, while maintaining comparable accuracies. This drastic performance improvement mainly results from the novel FL framework with forward-only propagation to achieve rapid convergence. LoLaFL is particularly beneficial when the data dimensionality and the class number are small, but low latency is required, especially in scenarios when computation and communication resources are severely limited and the data are non-IID.

Several directions for further research are worth exploring to overcome the limitation of LoLaFL. Firstly, the development of an improved coding theory to characterize the volume of the feature space and employing advanced optimization approaches could enhance the learning performance. Secondly, focusing on achieving higher compression rates, particularly in relation to exploiting the sparsity of the features, can further reduce the communication latency.
Thirdly, some dimensionality reduction techniques can be applied to the original data before they are utilized in LoLaFL. This approach can significantly reduce both communication latency and computation latency. Fourthly, when the number of classes is excessively large, it may be beneficial to train
multiple models using the LoLaFL scheme, with each model dedicated to handling a subset of classes. During inference, the predicted label for a given sample should
be determined by the class that receives the highest confidence score across the different models. Lastly, the impacts of the OFDMA-based channel allocation on LoLaFL could be further investigated.

\section*{Appendix}

\subsection*{A. Proof of Lemma 1}
For any permutation matrix $\mathbf{P}$, we have $\mathbf{Z}_\ell\mathbf{P}(\mathbf{Z}_\ell\mathbf{P})^*=\mathbf{Z}_\ell\mathbf{P}\mathbf{P}^*\mathbf{Z}_\ell^*\overset{(a)}{=}\mathbf{Z}_{\ell}\mathbf{Z}_{\ell}^*$, where $(a)$ is due to the property of the permutation matrix \cite{moon2000mathematical}. This suggests that the sample order of $\mathbf{Z}_\ell$ does not influence the global covariance matrix.
Therefore, without loss of generality, we let $\mathbf{Z}_\ell\triangleq\left[\mathbf{Z}_{\ell,1}, \ \mathbf{Z}_{\ell,2}, \ \dots, \ \mathbf{Z}_{\ell,K}\right]$, and thus
\begin{equation}
\setlength{\arraycolsep}{1.5pt}
\renewcommand{\arraystretch}{0.5}
\mathbf{\Pi}^j=\begin{bmatrix}
        \mathbf{\Pi}_1^{j} &  & & \\
         & \mathbf{\Pi}_2^{j} & & \\
         &  & \ddots  & \\
         &  &  & \mathbf{\Pi}_K^j
        \end{bmatrix}
.\label{Pi_j}
\end{equation}
By using matrix partition and the corresponding multiplication law, we have \eqref{global ZZ} and \eqref{global ZPZ}.
\begin{align}
\mathbf{Z}_{\ell}\mathbf{Z}_{\ell}^*&=\left[\mathbf{Z}_{\ell,1}, \ \mathbf{Z}_{\ell,2}, \ \dots, \ \mathbf{Z}_{\ell,K}\right]\left[\mathbf{Z}_{\ell,1}, \ \mathbf{Z}_{\ell,2}, \ \dots, \ \mathbf{Z}_{\ell,K}\right]^* \nonumber\\
&=\sum_{k=1}^K\mathbf{Z}_{\ell,k}\mathbf{Z}_{\ell,k}^*\label{global ZZ} 
\end{align}

\begin{align}
\mathbf{Z}_{\ell}\mathbf{\Pi}^j\mathbf{Z}_{\ell}^*
   =&\left[\mathbf{Z}_{\ell,1}, \ \mathbf{Z}_{\ell,2}, \ \dots, \ \mathbf{Z}_{\ell,K}\right]
   \setlength{\arraycolsep}{1.5pt}
  \renewcommand{\arraystretch}{0.5}
   \begin{bmatrix}
        \mathbf{\Pi}_1^{j} & & & \\
         & \mathbf{\Pi}_2^{j} & & \\
         & & \ddots & \\
         & & & \mathbf{\Pi}_{K}^j
        \end{bmatrix}
        \begin{bmatrix}
        \mathbf{Z}_{\ell,1}^{*}  \\
         \mathbf{Z}_{\ell,2}^{*}  \\
        \vdots  \\
         \mathbf{Z}_{\ell,K}^{*}
        \end{bmatrix}\nonumber\\
=&\sum_{k=1}^K\mathbf{Z}_{\ell,k}\mathbf{\Pi}_{k}^j\mathbf{Z}_{\ell,k}^{*}\label{global ZPZ} 
\end{align}

Therefore, the proof is completed.\qed

\subsection*{B. Proof of Proposition 1}

By transforming \eqref{El_k} and \eqref{Cl_jk}, we have \eqref{Z_l^kZ_l^k*} and \eqref{Z_l^kPiZ_l^k*}.
\begin{equation}
\mathbf{Z}_{\ell,k}\mathbf{Z}_{\ell,k}^{*}=({1}/{\alpha_k})(\mathbf{E}_{\ell,k})^{-1}-({1}/{\alpha_k})\mathbf{I}\label{Z_l^kZ_l^k*}
\end{equation}
\begin{equation}
\mathbf{Z}_{\ell,k}\mathbf{\Pi}_k^j\mathbf{Z}_{\ell,k}^{*}=({1}/{\alpha_k^j})(\mathbf{C}_{\ell,k})^{-1}-({1}/{\alpha_k^j})\mathbf{I}\label{Z_l^kPiZ_l^k*}
\end{equation}

Therefore, we have
\begin{equation}
    \label{proof_El}
    \begin{split}
    \Bar{\mathbf{E}}_\ell=&(\mathbf{I}+\alpha \mathbf{Z}_\ell\mathbf{Z}_\ell^*)^{-1}\\
        \overset{(a)}{=}&\left(\mathbf{I}+\alpha \sum_{k=1}^{K} \mathbf{Z}_{\ell,k}\mathbf{Z}_{\ell,k}^{*}\right)^{-1}\\
        \overset{(b)}{=}&\left(\mathbf{I}+\alpha \sum_{k= 1}^{K}\Bigl( ({1}/{\alpha_k})(\mathbf{E}_{\ell,k})^{-1}-({1}/{\alpha_k})\mathbf{I}\Bigl) \right)^{-1}\\
        =&\left(\Bigl(1-\alpha\sum_{k=1}^{K}{1}/{\alpha_k}\Bigl)\mathbf{I}+ \sum_{k= 1}^{K}\frac{\alpha}{\alpha_k} (\mathbf{E}_{\ell,k})^{-1}\right)^{-1}\\
        \overset{(c)}{=}&\left( \sum_{k= 1}^{K} \omega_k(\mathbf{E}_{\ell,k})^{-1}\right)^{-1},
    \end{split}
\end{equation}
where equality $(a)$ holds because of Lemma 1, $(b)$ holds because of \eqref{Z_l^kZ_l^k*}, and $(c)$ holds because $\sum_{k=1}^{K}{1}/{\alpha_k}=\sum_{k=1}^{K}{m_k \epsilon^2}/{d}={m \epsilon^2}/{d}={1}/{\alpha}$ and $\alpha/\alpha_k=({d}/{m \epsilon^2})/({d}/{m_k \epsilon^2})=m_k/m=\omega_k$.
Also, we have
\begin{equation}
    \label{proof_Cl_j}
    \begin{aligned}        \Bar{\mathbf{C}}_\ell^j=&(\mathbf{I}+\alpha^j \mathbf{Z}_\ell\mathbf{\Pi} ^j \mathbf{Z}_\ell^*)^{-1}\\
        \overset{(a)}{=}&\left(\mathbf{I}+\alpha^j \sum_{k=1}^{K} \mathbf{Z}_{\ell,k}\mathbf{\Pi}_k^j\mathbf{Z}_{\ell,k}^*\right)^{-1}\\
        \overset{(b)}{=}&\left(\mathbf{I}+\alpha^j \sum_{k=1}^{K}\Bigl( ({1}/{\alpha_k^j})(\mathbf{C}_{\ell,k}^{j})^{-1}-({1}/{\alpha_k^j})\mathbf{I}\Bigl) \right)^{-1}\\
        =&\left(\Bigl(1-\alpha^j\sum_{k=1}^{K}{1}/{\alpha_k^j}\Bigl)\mathbf{I}+ \sum_{k= 1}^{K} \frac{\alpha^j}{\alpha_k^j}(\mathbf{C}_{\ell,k}^{j})^{-1}\right)^{-1}\\
        \overset{(c)}{=}&\left(\sum_{k= 1}^{K}\omega_k^j (\mathbf{C}_{\ell,k}^{j})^{-1}\right)^{-1},\\
    \end{aligned}
\end{equation}
where equality $(a)$ holds because of Lemma 1, $(b)$ holds because of \eqref{Z_l^kPiZ_l^k*}, and $(c)$ holds because $\sum_{k=1}^{K}{1}/{\alpha_k^j}=\sum_{k=1}^{K}{\textrm{tr}(\mathbf{\Pi}_k^j) \epsilon^2}/{d}={\textrm{tr}(\mathbf{\Pi}^j) \epsilon^2}/{d}={1}/{\alpha^j}$ and $\alpha^j/\alpha_k^j=({d}/{\textrm{tr}(\mathbf{\Pi}^j) \epsilon^2})/({d}/{\textrm{tr}(\mathbf{\Pi}_k^j) \epsilon^2})=\textrm{tr}(\mathbf{\Pi}_k^j)/\textrm{tr}(\mathbf{\Pi}^j)=\omega_k^j$.\qed

\bibliographystyle{ieeetr}
\bibliography{Ref}

\end{document}